\documentclass[journal]{IEEEtran}
\ifCLASSINFOpdf

\else

\fi
\usepackage{amsmath}
\usepackage{amssymb}
\usepackage{booktabs}
\usepackage{color}
\usepackage{xcolor}
\usepackage{ulem}
\usepackage{cite}
\usepackage{bm}
\usepackage{graphicx}

\newcommand{\revision}[1]{\textcolor{black!72!black}{#1}}
\newcommand{\revisiontwo}[1]{\textcolor{black!72!black}{#1}}
\newcommand{\bmde}[1]{\revisiontwo{\bm{#1}}}

\newcommand{\RNum}[1]{\uppercase\expandafter{\romannumeral #1\relax}}
\begin{document}
\title{Three-Dimensional Dynamic Modeling and Motion Analysis for an Active-Tail-Actuated Robotic Fish with Barycentre Regulating Mechanism}
\author{Xingwen Zheng$^{1}$, Minglei Xiong$^{1,2}$, Junzheng Zheng$^{1}$, Manyi Wang$^{1}$, Runyu Tian$^{1,3}$, and Guangming Xie$^{1,4,5}$
\thanks{$^{1}$Xingwen Zheng, Minglei Xiong, Junzheng Zheng, Manyi Wang, Runyu Tian, and Guangming Xie are with the State Key Laboratory for Turbulence and Complex Systems, Intelligent Biomimetic Design Lab, College of Engineering, Peking University, Beijing, 100871, China.
        {\tt\small \{zhengxingwen, xiongml, zhengjunzheng, wangmanyi, trytian, xiegming\}@pku.edu.cn.} Corresponding author: G. Xie.}
\thanks{$^{2}$Minglei Xiong is with the Boya Gongdao (Beijing) Robot Technology Co., Ltd., Beijing, 100084, China.
}
\thanks{$^{3}$Runyu Tian is with the China Aerodynamics Research and Development Center, Mianyang, Sichuan, 621000, China.
}
\thanks{$^{4}$Peng Cheng Laboratory, Shenzhen, 518055, China.
}
\thanks{$^{5}$Guangming Xie is with the Institute of Ocean Research, Peking University, Beijing, 100871, China.
}}

\maketitle
\begin{abstract}
Dynamic modeling has been capturing attention for its fundamentality in precise locomotion analyses and control of underwater robots. However, the existing researches have mainly focused on investigating two-dimensional motion of underwater robots, \revisiontwo{and} little attention has been paid to three-dimensional dynamic modeling\revisiontwo{,} which is just what we focus on. In this article, a three-dimensional dynamic model of an active-tail-actuated robotic fish with \revisiontwo{a} barycentre regulating mechanism is built by combining Newton's second law for linear motion and Euler's equation for angular motion. The model parameters are determined by three-dimensional computer-aided design (CAD) software SolidWorks, HyperFlow-based computational fluid dynamics (CFD) simulation, and grey-box model estimation method. Both kinematic experiments \revision{with a prototype} and numerical simulations are applied to \revisiontwo{validate the accuracy of the dynamic model mutually}. Based on the dynamic model, multiple three-dimensional motions\revisiontwo{,} including rectilinear motion, turning motion, gliding motion, and spiral motion\revisiontwo{,} are analyzed. The experimental and simulation results demonstrate the effectiveness of the proposed model in evaluating the trajectory, attitude, and motion parameters\revisiontwo{,} including the velocity, turning radius, angular velocity, etc., of the robotic fish.
\end{abstract}

\begin{IEEEkeywords}
Three-dimensional dynamic modeling, Newton-Euler method, computational fluid dynamics (CFD), grey-box model estimation, robotic fish.
\end{IEEEkeywords}

\IEEEpeerreviewmaketitle

\section{Introduction}
\IEEEPARstart{I}n recent years, underwater robots including varieties of underwater remotely operated vehicles (ROV), autonomous underwater vehicles (AUV), and bio-inspired aquatic systems \cite{salazar2018classification} have been developed and shown great potentials in promoting marine resource exploitation \cite{khatib2016ocean,huang2018efficient}, marine economy development \cite{ryuh2015school,chang2016hunting}, and marine ecological environment protection \cite{wu2017development,vasilijevic2017coordinated}. The research topics of underwater robots \revision{cover} locomotion control and optimization \cite{shi2017advanced,li2017model}, underwater navigation and localization \cite{han2018matching,han2018combined}, environment perception and object recognition \cite{zheng2017artificial,aggarwal2015haptic}, underwater communication \cite{wang2017bio,marques2007auv}, etc. In particular, mechanism investigation of dynamic performance of underwater robots is fundamental and critical for the above-mentioned researches. \revisiontwo{Besides,} precise dynamic modeling of underwater robots has always been focus and difficulty in underwater robot research.

For dynamic modeling, the typical modeling methods include Lagrangian dynamics method, Newton-Euler method, Lighthill's elongated-body theory, Schiehlen method, etc. Basing on the Newton-Euler method, Y. Shi's group has built a dynamic model of an AUV, and then investigated dynamic \revisiontwo{model-based} trajectory tracking control of planar motions of the AUV \cite{shen2017modified,shen2018path,shen2018trajectory}, \revision{without consideration of three-dimensional motions.} J. Yu's group has formulated a robotic fish dynamics using Schiehlen method \cite{yu2008three} and Lagrangian dynamics method \cite{yu2013dynamic}. \revisiontwo{It} has been demonstrated that the proposed dynamic model is efficient for seeking backward swimming pattern of the robotic fish \cite{yu2013dynamic}. They have also proposed a data-driven dynamic modeling method in which the Newton-Euler formulation is applied to analyze the robotic fish dynamics, and parameters in the dynamic model are identified using experimental data of rectilinear motion and turning motion of the robotic fish, \revision{also without investigating three-dimensional motions.} F. Zhang's group has established an analytical model for spiral motion of an underwater glider steered by an internal movable mass block\revisiontwo{, and} experiments in the South China Sea \revisiontwo{have} validated the accuracy of the model for achieving desired spiral motion \cite{Zhang2013Spiraling}. They have also explored a dynamic model for a blade-driven glider with gliding motion \cite{Chen2016Design}. \revision{However, the motion of glider is different from \revisiontwo{rhythmic motion of the fin-actuated underwater robot}.} X. Tan's group has explored dynamic analyses of a tail-actuated robotic fish \cite{chen2010modeling,wang2013dynamic,wang2015averaging} and a fish-like glider \cite{zhang2014tail,zhang2014miniature}. For the tail-actuated robotic fish, Lighthill's large-amplitude elongated-body theory has been combined with rigid-body dynamics and hybrid tail dynamics for building a dynamic model \cite{chen2010modeling,wang2013dynamic,wang2015averaging}. \revision{However, only surface motion of the robotic fish has been explored.} For the fish-like glider, they have built a Newton-Euler method based dynamic model for investigating spiraling maneuver \cite{zhang2014tail} and gliding motion \cite{zhang2014miniature}. However, the fish-like glider is just driven by displacing an internal movable mass and pumping fluids, while its tail is not active, without a continuously varied tail angle.

The above-mentioned \revisiontwo{studies} have demonstrated that dynamic modeling is fundamental and essential for locomotion analysis of underwater robots. \revisiontwo{However, most of the researches have only focused on investigating two-dimensional motions in horizontal plane or vertical plane. Especially for fin-actuated underwater robots, though there are a few preliminary works \revisiontwo{that} have considered dynamic modeling in three-dimensional space \cite{yu2008three,yu2013dynamic,ozmen2018three}, the proposed models are typically validated by limited experiments, without validation in a large-scale parameter space. Besides, for three species of underwater robots including active-fin-actuated underwater robot with barycentre regulating mechanism, blade-driven underwater robot \cite{Chen2016Design}, and internal movable mass block-driven underwater robot \cite{wang2013dynamic}, all of which can adjust their centers of mass, there exist significant differences among their dynamics, because an active-fin-actuated underwater robot with barycentre regulating mechanism is able to generate extra rhythmic oscillation of robot body. However, dynamic modelling for such an underwater robot has been rarely investigated.}

On the basis of the above analyses, this article mainly focuses on investigating three-dimensional dynamic modeling in \revision{a} large-scale parameter space for an active-tail-actuated robotic fish with a barycentre regulating mechanism, which has been rarely investigated. \revision{Multiple} swimming patterns including rectilinear motion, turning motion, gliding motion, and spiral motion are investigated. Firstly, a \revisiontwo{mathematical} description of \revisiontwo{the} dynamic model is proposed basing on Newton-Euler method. Then multiple methods\revisiontwo{,} including SolidWorks software, computational fluid dynamics (CFD) simulation, and grey-box model estimation method\revisiontwo{,} are used for determining model parameters. Finally, \revision{numerical simulations and massive kinematic experiments with a robotic fish prototype} in a large-scale parameter space are applied to mutually validate the accuracy of the dynamic model in predicting key features\revisiontwo{,} including trajectory, attitude, velocity, etc., of the robotic fish.

The remainder of this article is organized as follows. Section $\rm\RNum{2}$ introduces the bio-inspired robotic fish. Section $\rm\RNum{3}$ establishes a Newton-Euler dynamic model for the robotic fish and determines the model parameters. Section $\rm\RNum{4}$ presents simulation and experiment results. Section $\rm\RNum{5}$ concludes this article with an outline of future work.
\section{The Robotic Fish}
\begin{figure}[htb]
\centering
\includegraphics[width=0.9\columnwidth]{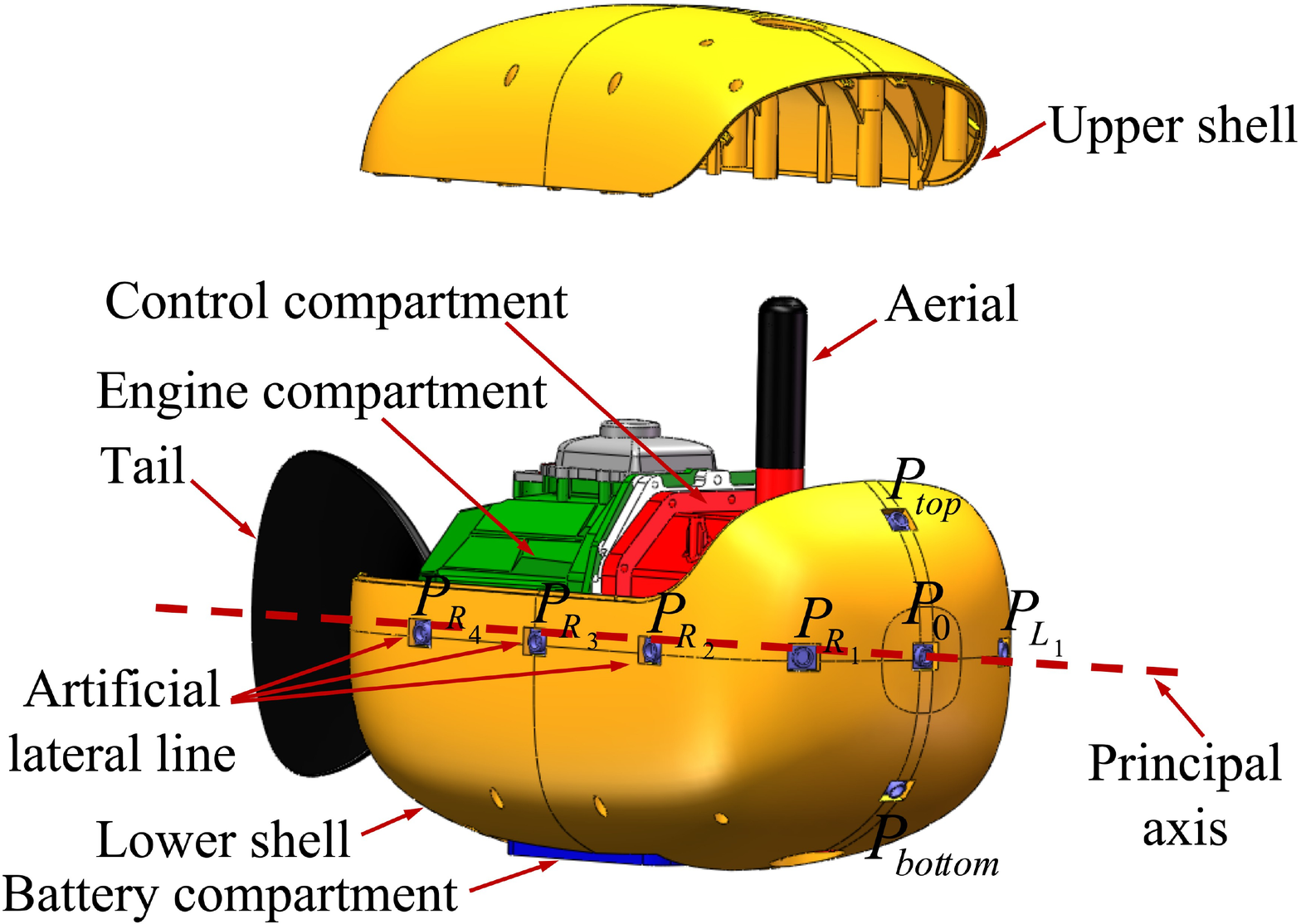}\\
(a)\\
\includegraphics[width=0.9\columnwidth]{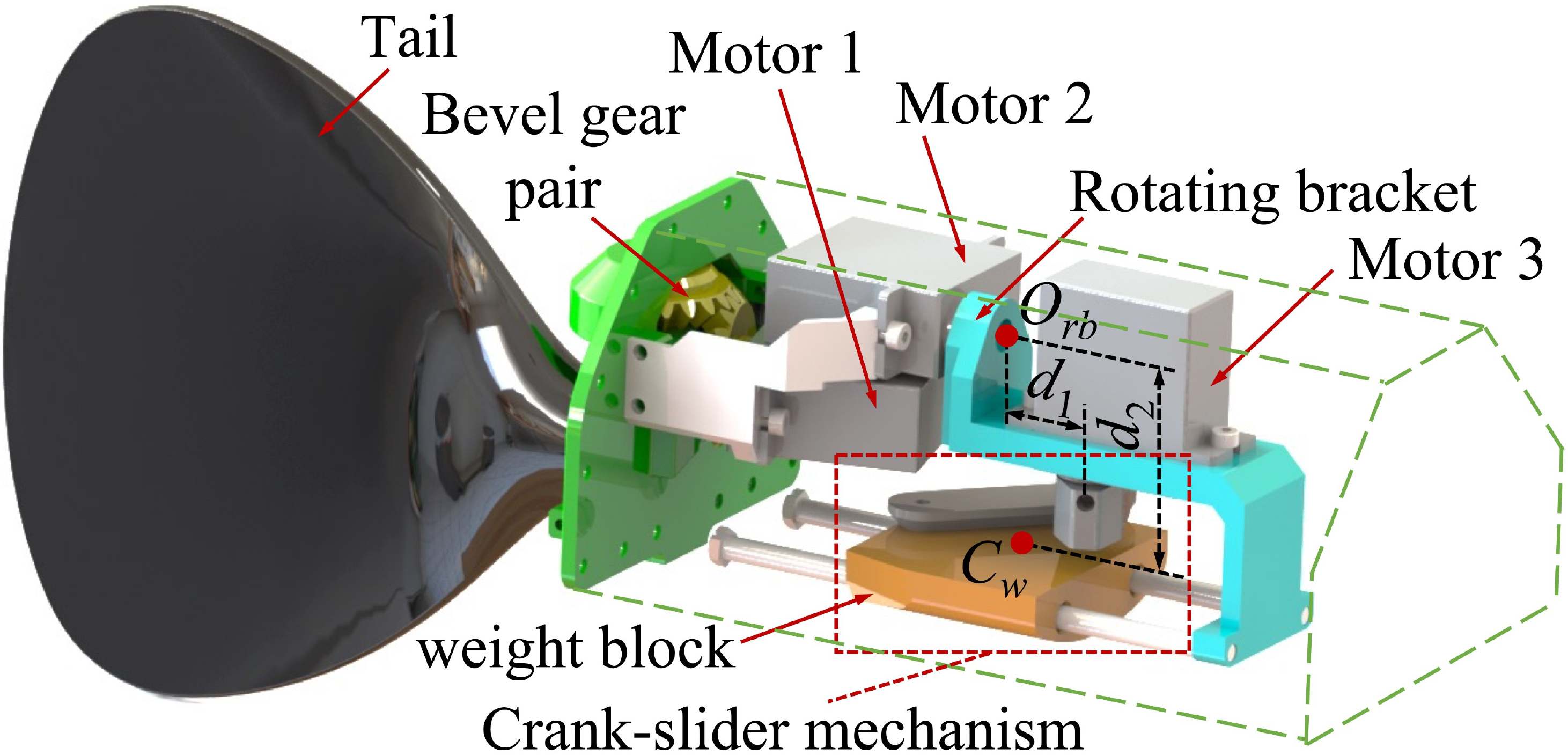}\\
\revision{(b)}
\caption{\revisiontwo{Hardware configurations of the robotic fish. (a) CAD model of the robotic fish. \revisiontwo{Eleven pressure sensors named $P_{top}$, $P_{bottom}$, $P_{0}$, $P_{L_i}$, and $P_{R_i}$ $(i=1,2,3,4)$ are mounted on the surface of the shell for establishing an artificial lateral line system (ALLS). ALLS is used to measure the hydrodynamic pressure variations surrounding fish body. More information about the ALLS can be found in our previous work \cite{zheng2017artificial}}. (b) The diagrammatic sketch of the interior of the engine compartment. $d_1$ indicates the distance between the output shaft of motor 3 and the connection point $O_{rb}$ of motor 2 and rotating bracket. $d_2$ indicates the distance between the output shaft of motor 2 and center of mass $C_w$ of the weight block.}}
\label{Hardware configurations of the robotic fish}
\end{figure}
\begin{figure*}[htb]
\centering
\includegraphics[width=\linewidth]{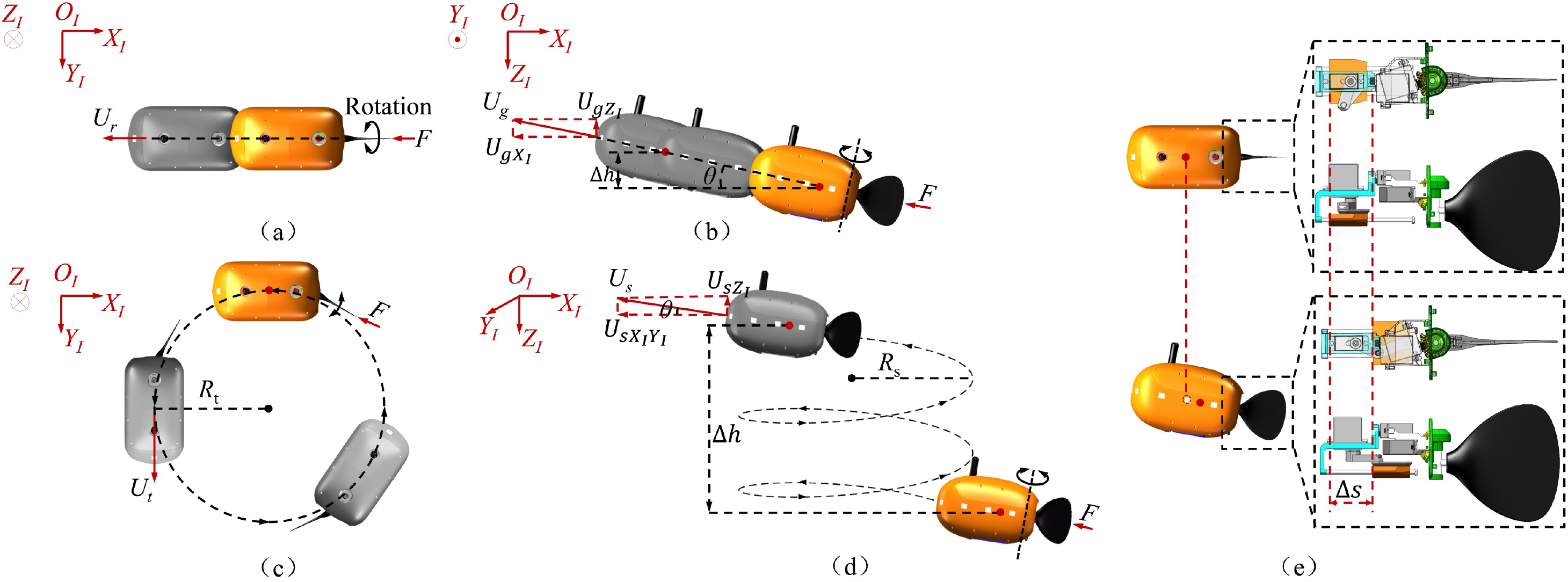}\\
\caption{\revisiontwo{Multiple three-dimensional swimming patterns of the robotic fish. (a) Rectilinear motion. (b) Gliding motion. (c) Turning motion. (d) Spiral motion. (e) The red point on fish shell means center of mass. It moves backward/forward when the weight block moves backward/forward with a distance of $\Delta s$ in gliding motion and spiral motion (lower), comparing with rectilinear motion and turning motion (upper). The tails in turning motion and spiral motion have non-zero offsets compared to those in rectilinear motion and gliding motion. $O_IX_IY_IZ_I$ indicates the global inertial coordinate system. $F$ indicates the tailed-generated propulsive force. $U_k (k=r, t, g, s)$ indicates the movement velocity of the robotic fish. $U_g$ is the resultant velocity of the velocity $V_{gZ_I}$ along the axis $O_IZ_I$ and the velocity $V_{gX_I}$ along the axis $O_IX_I$. $U_s$ is the resultant velocity of the velocity $V_{sZ_I}$ along the axis $O_IZ_I$ and the velocity $V_{sX_IY_I}$ on $X_I-Y_I$ plane. $R_t$ and $R_s$ indicates the radius in turning motion and spiral motion, respectively. $\Delta h$ indicates depth variation of the robotic fish. $\theta$ indicates pitch angle of the robotic fish.}}
\label{Multiple three dimensional swimming patterns of the robotic fish}
\end{figure*}
\revisiontwo{Figure~\ref{Hardware configurations of the robotic fish} (a) shows the hardware configurations of the robotic fish. Its size (Length$\times$Width$\times$Height) is about 29.1 cm$\times$11.6 cm$\times$13.4 cm. It is composed of a 3D-printed shell, a tail, and three compartments, including a control compartment, an engine compartment, a battery compartment, and a pressure acquisition system compartment. Figure~\ref{Hardware configurations of the robotic fish} (b) shows the interior of the engine compartment. Three motors\revisiontwo{,} which serve different functions\revisiontwo{,} are wrapped in the engine compartment. Specifically, motor 1 is connected with the tail. It is used to generate propulsive force. Motor 2 is used for drivinng a rotating bracket. The bracket is connected to motor 3 and a crank-slider mechanism. Motor 3 is used to drive the crank-slider mechanism mentioned above to which a weight block is connected. Through controlling motor 2 and 3, the weight block can move along the direction parallel to principal axis of the robotic fish and rotate about output shaft of motor 2. By controlling the three motors using given frequency, amplitude\revisiontwo{,} and offset parameters, the robotic fish can realize rectilinear motion, turning motion, gliding motion, and spiral motion, as shown in Figure~\ref{Multiple three dimensional swimming patterns of the robotic fish}. More about motions of the robotic fish can be in the supplementary video.}
\section{Dynamic Analysis for the Robotic Fish}
\subsection{Definition of the Coordinate Systems}
\begin{figure}[htb]
\centering
\includegraphics[width=0.9\columnwidth]{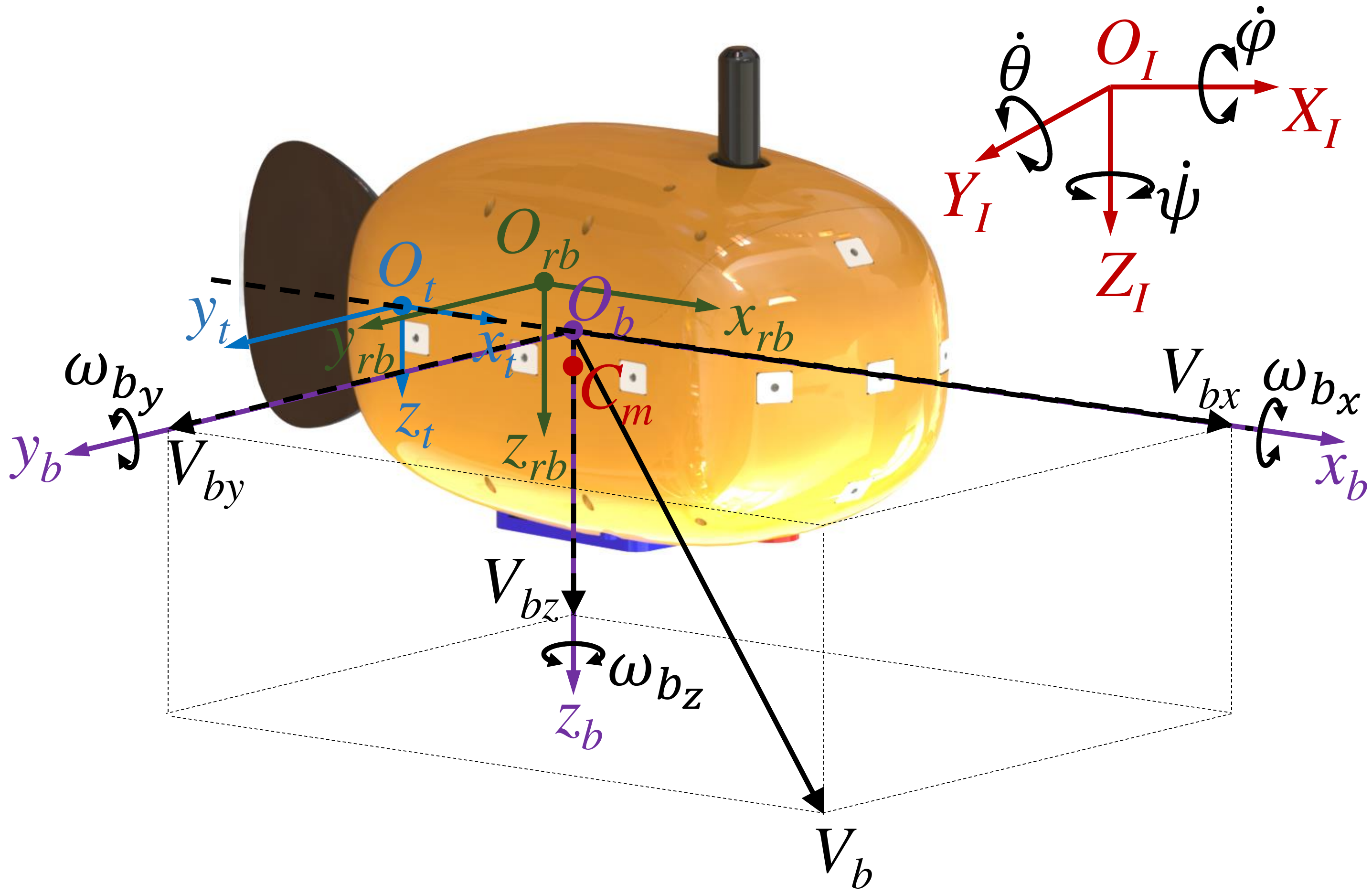}
\caption{\revisiontwo{Definitions of coordinate systems of the robotic fish.}}
\label{Definition of coordinate systems for the robotic fish}
\end{figure}

Figure~\ref{Definition of coordinate systems for the robotic fish} shows the coordinate systems of the robotic fish. $O_Ix_Iy_Iz_I$, $O_bx_by_bz_b$, $O_{rb}x_{rb}y_{rb}z_{rb}$, and $O_tx_ty_tz_t$ indicate the global inertial coordinate system, the body-fixed coordinate system, the rotating-bracket-fixed coordinate system, and the tail-fixed coordinate system, respectively. The origin $O_b$ is fixed at the intersection of horizontal section and longitudinal section of the robotic fish, above center of mass $C_m$ of the robotic fish. \revision{The longitudinal section is the symmetrical plane of the shell. The horizontal section coincides with the symmetrical plane of the tail and is perpendicular to the longitudinal plane.} The origin $O_{rb}$ is fixed at the connection point of \revision{motor} 2 and the rotating bracket in Figure~\ref{Hardware configurations of the robotic fish} (c), and expressed as $\left [ a_{rb}, b_{rb}, c_{rb} \right ]$ in $O_bx_by_bz_b$. The origin $O_t$ is fixed at the connection point of the tail and the engine compartment, and expressed as $\left [ a_t, b_t, c_t \right ]$ in $O_bx_by_bz_b$.
$O_Ix_Iy_Iz_I$ coincides with the initial $O_bx_by_bz_b$.
\subsection{Three-Dimensional Kinematic Analysis}
\subsubsection{Translational Motion of the Robotic Fish}
The position of the robotic fish is denoted as $\bmde{\revisiontwo{C_I}}=\left [ x_I, y_I, z_I \right ]^T$ in $O_IX_IY_IZ_I$. The velocity of robotic fish is denoted as $\bmde{V_I}=\left [ V_{Ix}, V_{Iy}, V_{Iz} \right ]^T$ in $O_IX_IY_IZ_I$ and $\bmde{V_b}=\left [ V_{bx}, V_{by}, V_{bz} \right ]^T$ in $O_bx_by_bz_b$, respectively. The relationship between $\bmde{V_I}$ and $\bmde{V_b}$ is expressed as
\begin{eqnarray}
\bmde{V_I}=\dot{\bmde{C_I}}=\bmde{R_{bI}}\cdot \bmde{V_b}
\label{velocity in the global inertial coordinate system}
\end{eqnarray}
where $\bmde{R_{bI}}$ is the transformation matrix from $O_bx_by_bz_b$ to $O_Ix_Iy_Iz_I$, taking the form as
\begin{eqnarray}
\bmde{R_{bI}}=\begin{bmatrix}
c_\psi c_\theta   &-s_\psi c_\varphi +c_\psi s_\theta s_\varphi   & s_\psi s_\varphi +c_\psi s_\theta c_\varphi \\
s_\psi c_\theta  &c_\psi c_\varphi +s_\psi s_\theta s_\varphi  & -c_\psi s_\varphi +s_\psi s_\theta c_\varphi \\
-s_\theta  & c_\theta s_\varphi  & c_\theta c_\varphi
\end{bmatrix}
\label{transformation matrix from the body-fixed coordinate system to the global inertial coordinate system}
\end{eqnarray}
where $\varphi$, $\theta$, and $\psi$ indicate roll, pitch, and yaw angle of the robotic fish, respectively.
\subsubsection{Rotational Motion of the Robotic Fish}
The angular velocity of the robotic fish is denoted as $\bmde{\omega_b}=\left [ \omega_{b_x}, \omega_{b_y}, \omega_{b_z} \right ]^T$ in $O_bx_by_bz_b$ and $\bmde{\omega_I}=\left [ \dot{\varphi }, \dot{\theta }, \dot{\psi } \right ]^T$ in $O_Ix_Iy_Iz_I$. The relationship between $\bmde{\omega_b}$ and $\bmde{\omega_I}$ is expressed as
\setlength{\arraycolsep}{0.0em}
\begin{eqnarray}
\bmde{\omega_I}=\setlength{\arraycolsep}{5pt}\begin{bmatrix}
1 & sin\varphi tan\theta & cos\varphi tan\theta \\
0 & cos\varphi  & -sin\varphi  \\
0 & sin\varphi/cos\theta  & cos\varphi/cos\theta
\end{bmatrix}\cdot\bmde{\omega_b}
\label{kinematics equation 2}
\end{eqnarray}
\setlength{\arraycolsep}{5pt}
\subsubsection{Motion analysis of the weight block}
\begin{figure}[htb]
\centering
\includegraphics[width=0.83\linewidth]{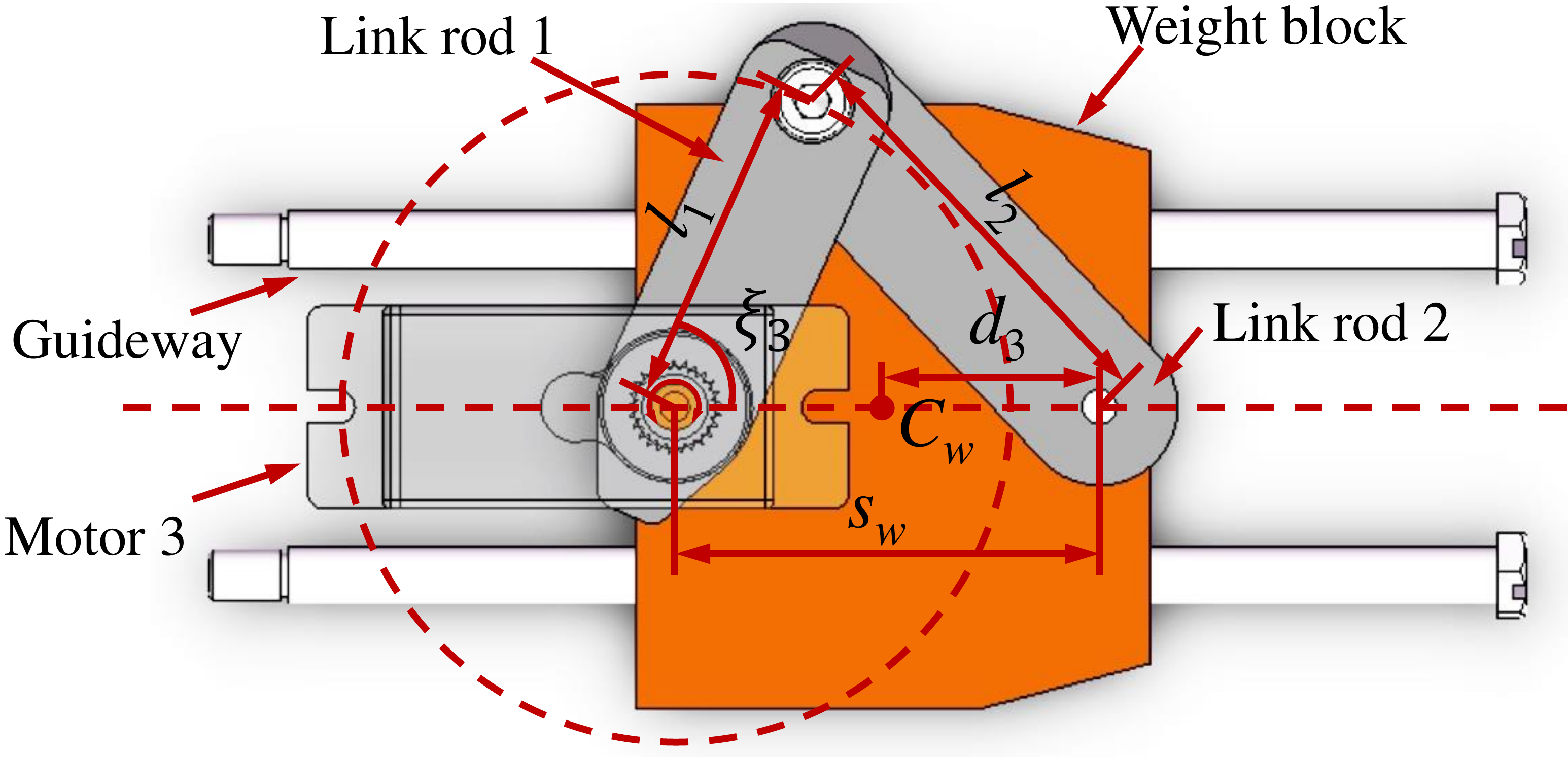}\\
\caption{\revisiontwo{Crank-slider mechanism with the weight block. $l_1$ and $l_2$ indicate the lengths of the link rods. $d_3$ indicates the distance between center of mass of the weight block $C_w$ and connecting point of the weight block and link rod 2. $s_w$ indicates the distance between output shaft of motor 3 and connecting point of the weight block and link rod 2. The masses of guideway, motor 3, link rod 1, and link 2 are all ignored.}}
\label{Crank-slider mechanism with weight block}
\end{figure}

As shown in Figure~\ref{Hardware configurations of the robotic fish} (c), the weight block is able to rotate through controlling output angle $\xi_2$ of \revision{motor} 2. Thus roll angle $\varphi$ of the robotic fish is able to be adjusted. On the other hand, as shown in Figure~\ref{Crank-slider mechanism with weight block}, the distance $s_w$ is able to be adjusted through controlling output angle $\xi_3$ of \revision{motor} 3. Thus the weight block is able to move along the guideway\revision{,} and pitch angle $\theta$ of the robotic fish is able to be adjusted. $s_w$ takes the form as
\begin{eqnarray}
s_{w}=s_{w_0}+\Delta d
\label{distance used for adjusting the pitch angle}
\end{eqnarray}
where $s_{w_0}$ indicates the initial value of $s_w$, with which pitch angle and roll angle of the robotic fish \revision{are} 0. $\Delta d$ is the distance between the weight block's current position and its initial position in the kinematics experiments.

The coordinate of center of mass of the weight block $C_w[x_{C_w}, y_{C_w}, z_{C_w}]$ is expressed in $O_bx_by_bz_b$, taking the form as
\begin{eqnarray}
\begin {aligned}
x_{C_{w}}&=a_{rb}+d_{1}-\left (s_{w}-d_{3} \right )\\
y_{C_{w}}&=b_{rb}+d_{2}\cdot sin\xi_{2}\\
z_{C_{w}}&=c_{rb}+d_{2}\cdot cos\xi_{2}
\end {aligned}
\end{eqnarray}

The coordinate of center of mass of the robotic fish $C_m[x_{C_m}, y_{C_m}, z_{C_m}]$ takes the form as
\begin{eqnarray}
\begin {aligned}
j_{C_m}=\frac{\left ( M_{ew_{j}}+M_{w_{j}} \right )}{m_{total}}
\end{aligned}
\label{coordinates of center of mass of the robotic fish}
\end{eqnarray}
where $j=x, y, z$, $M_{ew_{j}}$ is static moment about the $O_bj_b$ axis for the part apart from the weight block. $M_{w_{j}}$ is static moment about the $O_bj_b$ axis for the weight block, taking the form as
\begin{eqnarray}
\begin {aligned}
M_{w_{j}}=m_{w}\cdot j_{C_{w}}
\end {aligned}
\label{static moments of the weight block}
\end{eqnarray}
where $m_w$ is the mass of the weight block.

The initial coordinate of center of mass of the robotic fish is expressed as $[x_{C_{m0}}, y_{C_{m0}}, z_{C_{m0}}]$. \revisiontwo{Besides,} both pitch angle $\theta$ and roll angle $\varphi$ of the robotic fish are zero when the weight block is at its initial position.
\subsection{Three-Dimensional Force Analysis}
In this part, the forces and torques acting on the tail and fish body of the robotic fish are analyzed. For the tail, the lift and drag are considered. For the fish body, we respectively consider lift force and drag force in $x_b-z_b$ plane and $x_b-y_b$ plane, \revision{gravity, buoyancy}, and impact of water flow.
\subsubsection{Force Analysis for the Tail}
\begin{figure}[htb]
\centering
\includegraphics[width=0.65\columnwidth]{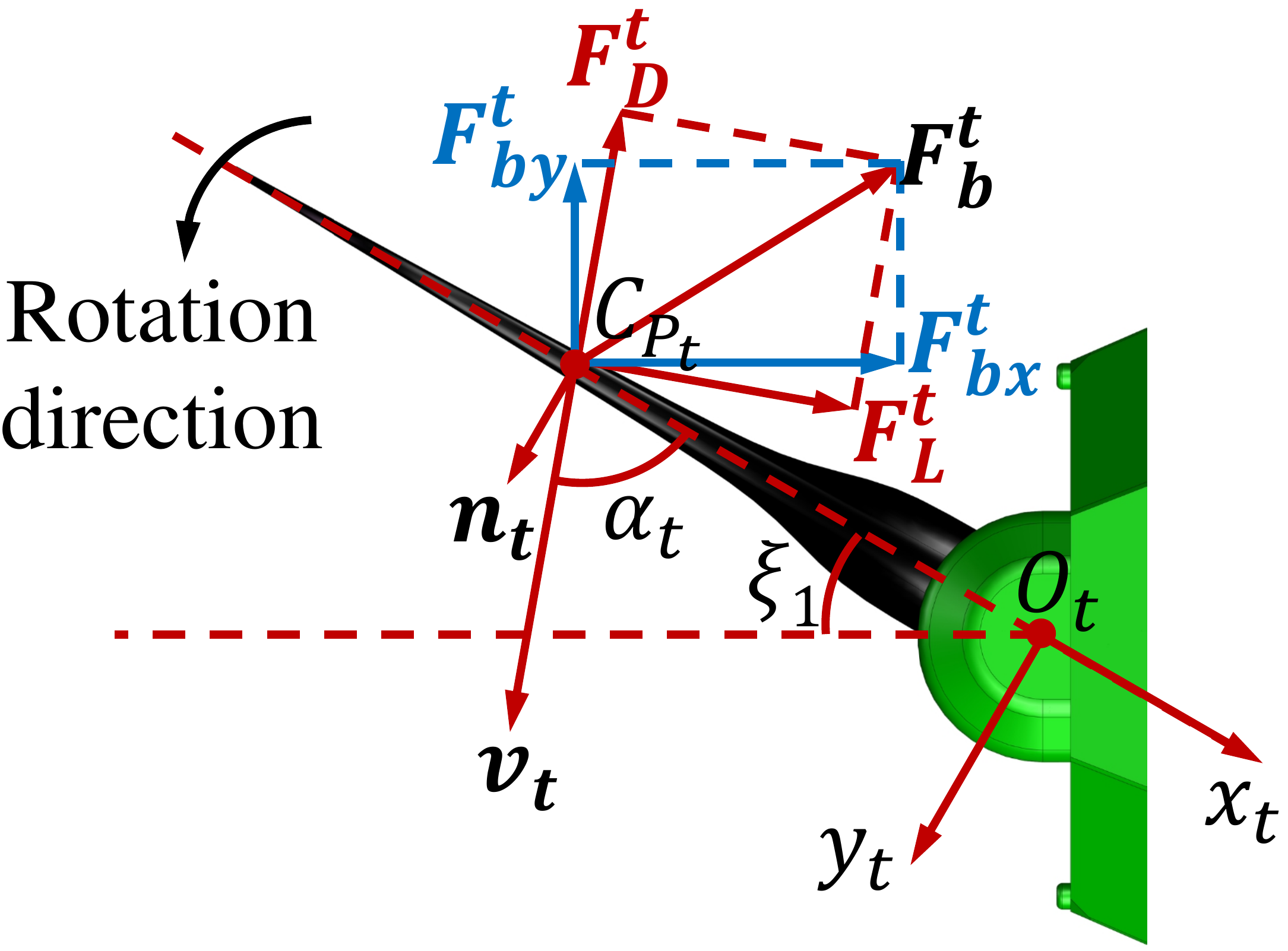}
\caption{\revisiontwo{Force analysis for the tail. $C_{p_t}$ is center of press of the tail, and it is coincident with center of mass of the tail.}}
\label{Force analysis for the tail}
\end{figure}
For the tail of the robotic fish, its time-varying oscillating angle $\xi_1$ is expressed as
\begin{eqnarray}
\xi_{1} \left ( t \right )=\bar{\xi_{1}}+A_{1}sin\left ( 2\pi f_{1}t \right )
\label{oscillating angle of the tail}
\end{eqnarray}
where $\bar{\xi_{1}}$, $A_{1}$, and $f_{1}$ are the oscillating offset, amplitude, and frequency of the tail, respectively.

The velocity of $C_{p_t}$ in Figure~\ref{Force analysis for the tail} is expressed as
\begin{eqnarray}
\bmde{v_{t}}=\bmde{V_b}+\bmde{\omega_b}\times \bmde{O_bC_{P_t}}+\bmde{\omega_t}\times \bmde{O_tC_{P_t}}
\label{velocity of center of press (C. P.) of the tail}
\end{eqnarray}
where $\bmde{O_bC_{p_t}}$ is the vector from $O_b$ to $C_{p_t}$. \revisiontwo{It} is expressed as
\setlength{\arraycolsep}{0.0em}
\begin{eqnarray}
\bmde{O_bC_{p_t}}&=&\left ( a_{t}\!-\!r_{c}\!\cdot\! cos\xi _{1}  \right )\!\cdot\!\bmde{\hat{x_b}}\!+\!(b_{t}\!-\!r_{c}\!\cdot\! sin\xi _{1})\!\cdot\!\bmde{\hat{y_b}}\!+\!c_{t}\!\cdot\!\bmde{\hat{z_b}}
\label{vector from $O_b$ to center of mass of the tail}
\end{eqnarray}
where $\bmde{\hat{x_{b}}}$, $\bmde{\hat{y_{b}}}$, and $\bmde{\hat{z_{b}}}$ are unit vector along the $O_bx_b$ axis, $O_by_b$ axis, and $O_bz_b$ axis in $O_bx_by_bz_b$, respectively.
$\bmde{O_tC_{p_t}}$ is the vector from $O_t$ to $C_{p_t}$, and it is expressed as
\setlength{\arraycolsep}{5pt}
\begin{eqnarray}
\bmde{O_tC_{p_t}}=-r_c\cdot cos\xi_{1}\cdot\bmde{\hat{x_b}}-r_c\cdot sin\xi_{1}\cdot\bmde{\hat{y_b}}+0\cdot\bmde{\hat{z_b}}
\label{vector from $O_t$ to center of press of the tail}
\end{eqnarray}
$\bmde{\omega_{t}}$ is the oscillating angular velocity of the tail, and it is expressed as
\begin{eqnarray}
\bmde{\omega_{t}}=\dot{\xi_{1}}\cdot\bmde{\hat{z_b}}=2\pi f_{1}A_{1}cos\left ( 2\pi f_{1}t \right )\cdot\bmde{\hat{z_b}}
\label{oscillating angular velocity of the tail}
\end{eqnarray}

\revision{The tail of the robotic fish is regarded as a rigid plate without spanwise wave motion, which is different from fins in \cite{tangorra2009biorobotic}. There are various forms of tail-generated force and torque \cite{yun2014novel,Morgansen2007Geometric,chen2013target,wang2015averaging,ozmen2018three,hu2009vision} for different of types of tails. Here, we have adopted forms as in \cite{wang2015averaging,ozmen2018three,hu2009vision}, which are typically applied to express torque and force caused by a rigid plate-like tail. \revisiontwo{Specifically,} the lift $F_{L}^{t}$ and drag $F_{D}^{t}$ of the tail are expressed as}
\begin{eqnarray}
\begin {aligned}
F_{\lambda}^{t}&=\frac{1}{2}\rho \left | \bmde{v_{t}} \right |^{2}S_{t}C_{\lambda_{t}}(\left |\alpha_{t}\right |)\\
\end {aligned}
\label{The lift and drag of the tail}
\end{eqnarray}
where $\lambda=L, D$. $\rho$ is the density of water. $S_t$ is the surface area of the tail. $C_{L_{t}}$ and $C_{L_{t}}$ are force coefficients which will be determined in section $\rm\RNum{3}$. E. $\alpha _{t}$ is the angle of attack of the tail, which is expressed as
\begin{eqnarray}
\alpha _{t}=arcsin\left ( \bmde{n_{t}}\cdot \bmde{\hat{v_{t}}} \right )
\label{angle of attack of the tail}
\end{eqnarray}
where $\bmde{n_{t}}$ is the normal vector of the tail, which is expressed as
\begin{eqnarray}
\bmde{n_{t}}=-sin\xi_{1}\cdot\bmde{\hat{x_b}}+cos\xi_{1}\cdot\bmde{\hat{y_b}}+0\cdot\bmde{\hat{z_b}}
\label{normal vector of the tail}
\end{eqnarray}

Basing on the above analyses, the three-dimensional drag $\bmde{F_{D}^{t}}$ \cite{Krause2005Fluid} acting on the tail is expressed as
\begin{eqnarray}
\bmde{F_{D}^{t}}=-F_{D}^{t}\bmde{\hat{v_{t}}}
\label{The drag acting on the tail}
\end{eqnarray}

The three-dimensional lift $\bmde{F_{L}^{t}}$ acting on the tail is expressed as
\begin{eqnarray}
\bmde{F_{L}^{t}}=\left\{\begin{matrix}
\frac{\bmde{v_{t}}sin\alpha_t-\bmde{n_{t}}}{\left \| \bmde{v_{t}}sin\alpha_t-\bmde{n_{t}} \right \|}\cdot F_{L}^{t} & \text{ if } \bmde{n_{t}}\cdot \bmde{\hat{v_{t}}}> 0\\
\frac{\bmde{v_{t}}sin\alpha_t+\bmde{n_{t}}}{\left \| \bmde{v_{t}}sin\alpha_t+\bmde{n_{t}} \right \|}\cdot F_{L}^{t} & \text{ if } \bmde{n_{t}}\cdot \bmde{\hat{v_{t}}}\leq 0
\end{matrix}\right.
\label{The lift acting on the tail}
\end{eqnarray}

Then, the tail-generated torque $\bmde{M_{b}^{t}}$ acting on the robotic fish is expressed as
\begin{eqnarray}
\bmde{M_{b}^{t}}=\bmde{O_bC_{p_t}}\times (\bmde{F_{L}^{t}}+\bmde{F_{D}^{t}})
\label{The tail-generated torque acting on the robotic fish}
\end{eqnarray}
\subsubsection{Force Analysis for Fish Body}
\begin{figure}[htb]
\centering
\includegraphics[width=0.65\columnwidth]{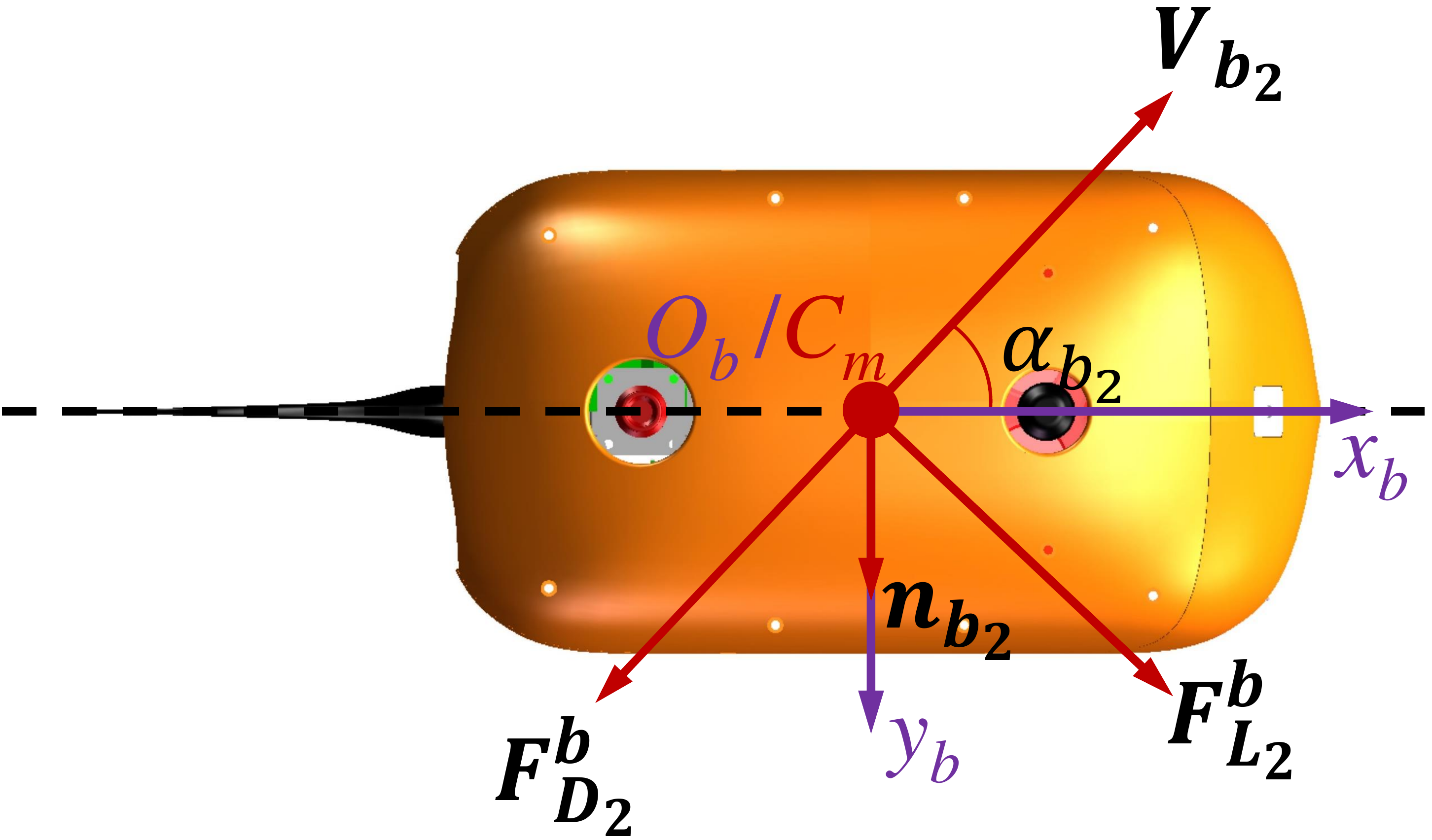}\\
(a)\\
\includegraphics[width=0.65\columnwidth]{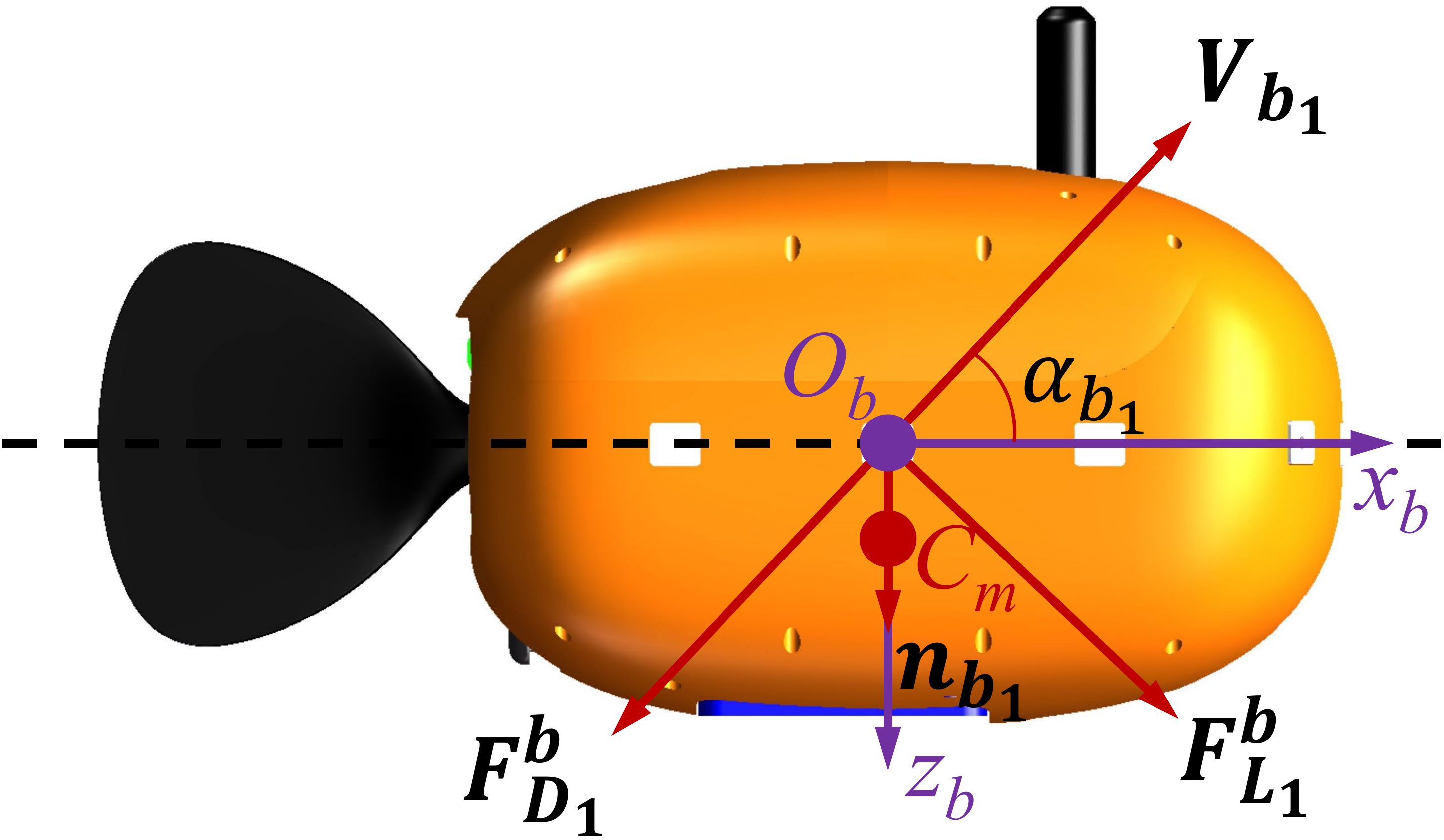}\\
(b)
\caption{\revisiontwo{Force analysis for the fish body. (a) Force analysis for $x_b-z_b$ plane. (b) Force analysis for $x_b-y_b$ plane.}}
\label{Force analysis for the fish body}
\end{figure}

Figure~\ref{Force analysis for the fish body} shows the drag $\bmde{F_{D_i}^{b}} (i=1, 2)$ and lift $\bmde{F_{L_i}^{b}} (i=1, 2)$ acting on the fish body, of which the values are expressed as
\begin{eqnarray}
\begin {aligned}
F_{D_i}^{b}&=\frac{1}{2}\rho \left | \bmde{V_{b_i}} \right |^{2}S_{b_i}C_{D_{b_i}}(\left |\alpha_{b_i}\right |)\\
F_{L_i}^{b}&=\frac{1}{2}\rho \left | \bmde{V_{b_i}} \right |^{2}S_{b_i}C_{L_{b_i}}(\left |\alpha_{b_i}\right |)
\end {aligned}
\end{eqnarray}
where $C_{D_{b_i}}$ and $C_{L_{b_i}}$ are force coefficients which will be determined in section$\rm\RNum{3}$. E.
\begin{eqnarray}
\begin {aligned}
\bmde{V_{b_1}}&=V_{bx}\cdot\bmde{\hat{x_b}}+V_{bz}\cdot\bmde{\hat{z_b}}\\
\bmde{V_{b_2}}&=V_{bx}\cdot\bmde{\hat{x_b}}+V_{by}\cdot\bmde{\hat{y_b}}
\end {aligned}
\end{eqnarray}
$S_{b_i} (i=1, 2)$ is the surface area tensor of the robotic fish. \revisiontwo{It} is defined as
\begin{eqnarray}
S_{b_i}=\bmde{\hat{V_{b_i}}}^{T}\cdot \bmde{A_i}\cdot \bmde{\hat{V_{b_i}}}, (i=1, 2)
\end{eqnarray}
where
\begin{eqnarray}
\bmde{A_1}=\begin{bmatrix}
S_{xx} & S_{xz}\\
S_{zx} & S_{zz}
\end{bmatrix}, \bmde{A_2}=\begin{bmatrix}
S_{xx} & S_{xy}\\
S_{yx} & S_{yy}
\end{bmatrix}
\end{eqnarray}
$\bmde{A_1}$ and $\bmde{A_2}$ are diagonal matrices. $S_{xx}$, $S_{yy}$, and $S_{zz}$ indicates the maximum cross section area perpendicular to the axes $O_bx_b$, $O_by_b$, and $O_bz_b$. $\alpha _{b_i} (i=1, 2)$ is angle of attack of fish body, \revision{taking the form} as
\begin{eqnarray}
\alpha _{b_i}=arcsin\left ( \bmde{n_{b_i}}\cdot \bmde{\hat{V_{b_i}}} \right )
\end{eqnarray}
$\bmde{n_{b_i}} (i=1, 2)$ is the normal vector, taking the form as
\begin{eqnarray}
\bmde{n_{b_1}}=\bmde{\hat{z_b}}, \bmde{n_{b_2}}=\bmde{\hat{y_b}}
\end{eqnarray}

Basing on the above-analyses, the three-dimensional drag $\bmde{F_{D_i}^{b}} (i=1, 2)$ is expressed as
\begin{eqnarray}
\bmde{F_{D_i}^{b}}=-F_{D_i}^{b}\bmde{\hat{V_{b_i}}}
\end{eqnarray}

The three-dimensional lift $\bmde{F_{L_i}^{b}} (i=1, 2)$ is expressed as
\begin{eqnarray}
\bmde{F_{L_i}^{b}}=\left\{\begin{matrix}
\frac{\bmde{V_{b_i}}sin\alpha_{b_i}-\bmde{n_{b_i}}}{\left \| \bmde{V_{b_i}}sin\alpha_{b_i}-\bmde{n_{b_i}} \right \|}\cdot F_{L_i}^{b} & \text{ if } \bmde{n_{b_i}}\cdot \bmde{\hat{V_{b_i}}}> 0\\
\frac{\bmde{V_{b_i}}sin\alpha_{b_i}+\bmde{n_{b_i}}}{\left \| \bmde{V_{b_i}}sin\alpha_{b_i}+\bmde{n_{b_i}} \right \|}\cdot F_{L_i}^{b} & \text{ if } \bmde{n_{b_i}}\cdot \bmde{\hat{V_{b_i}}}\leq 0
\end{matrix}\right.
\end{eqnarray}

Besides, rotations of the robotic fish cause damping torques $\bmde{M_{\omega}}$ acting on fish body\revisiontwo{, and} $\bmde{M_{\omega}}$ is expressed as
\setlength{\arraycolsep}{0.0em}
\begin{eqnarray}
\bmde{M_{\omega}}=\bmde{C_{\omega_b}}\cdot\bmde{\omega_{b}}
\label{drag torque caused by body rotation}
\end{eqnarray}
where $C_{\omega_b}$ is damping torque coefficient, taking the form as
\begin{eqnarray}
\bmde{C_{\omega_b}}=diag\left \{ C_{\omega_{b1}}, C_{\omega_{b2}}, C_{\omega_{b3}} \right \}
\label{damping coefficient}
\end{eqnarray}

In addition, the robotic fish is subjected to torque $\bmde{M_{I}}$ caused by impact of water flow\revisiontwo{, and} $\bmde{M_{I}}$ \cite{Krause2005Fluid} is expressed as
\begin{eqnarray}
\bmde{M_I}=M_{I_{x_b}}\cdot\bmde{\hat{x_b}}+M_{I_{y_b}}\cdot\bmde{\hat{y_b}}+M_{I_{z_b}}\cdot\bmde{\hat{z_b}}
\end{eqnarray}
where
\begin{eqnarray}
\begin {aligned}
M_{I_{x_b}}&=0\\
M_{I_{y_b}}&=\frac{1}{2}\rho \left | \bmde{V_{b_1}} \right |^{2}S_{b_1} C_{M_{I_{y_b}}}\left ( \alpha _{b_1} \right )\\
M_{I_{z_b}}&=\frac{1}{2}\rho \left | \bmde{V_{b_2}} \right |^{2}S_{b_2} C_{M_{I_{z_b}}}\left ( \alpha _{b_2} \right )
\end {aligned}
\end{eqnarray}
$C_{M_{I_{y_b}}}$ and $C_{M_{I_{z_b}}}$ are torque coefficients which will be determined in section $\rm\RNum{3}$. E.

\subsubsection{The Effect of Gravity and Buoyance}
The gravity $\bmde{F_{g}}$ and \revisiontwo{buoyancy} $\bmde{F_{b}}$ of the robotic fish are expressed in $O_bx_by_bz_b$, taking the form as
\begin{eqnarray}
\bmde{F_{g}}=m_{total}\cdot\bmde{R_{bI}}^{-1}\cdot\bmde{g}
\label{The gravity of the robotic fish}
\end{eqnarray}
\begin{eqnarray}
\bmde{F_{b}}=-m_{b}\cdot\bmde{R_{bI}}^{-1}\cdot\bmde{g}
\label{The buoyance of the robotic fish}
\end{eqnarray}
where $m_{total}$ and $m_b$ are total mass and buoyancy mass of the robotic fish, respectively.

The torque $\bmde{M_g}$ caused by the buoyance of the robotic fish is expressed as
\begin{eqnarray}
\bmde{M_g}=\bmde{O_{b}C_{m}}\times\bmde{F_{g}}
\label{The torque caused by the buoyance of the robotic fish}
\end{eqnarray}
where $\bmde{O_{b}C_{m}}$ is the vector from $O_b$ to $C_{m}$, taking the form as
\begin{eqnarray}
\bmde{O_{b}C_{m}}=x_{C_m}\bmde{\hat{x_b}}+y_{C_m}\bmde{\hat{y_b}}+z_{C_m}\bmde{\hat{z_b}}
\label{the vector from center of mass $C_{m}$ of the robotic fish to $O_b$}
\end{eqnarray}
\subsection{Newton-Euler Dynamic Model}
Basing on Newton's second law, the total force $\bmde{F_{total}}$ acting on the robotic fish is expressed as
\begin{eqnarray}
\left\{\begin{matrix}
\bmde{F_{total}}=\frac{d\bmde{M}\bmde{V_{C_m}}}{dt}\\
\bmde{F_{total}}=\bmde{F_g}\!+\!\bmde{F_{b}}\!+\!\bmde{F_{L_1}^{b}}\!+\!\bmde{F_{D_1}^{b}}\!+\!\bmde{F_{L_2}^{b}}\!+\!\bmde{F_{D_2}^{b}}\!+\!\bmde{F_{L}^{t}}\!+\!\bmde{F_{D}^{t}}
\end{matrix}\right.
\label{the total force acting on the robotic fish}
\end{eqnarray}
where $\bmde{M}=diag\left \{ m_{total}, m_{total}, m_{total} \right \}$. $\bmde{V_{C_m}}$ indicates velocity of  center of mass $C_m$ of the robotic fish, taking the form as
\begin{eqnarray}
\bmde{V_{C_m}}=\bmde{V_b}+\bmde{\omega_b}\times\bmde{O_{b}C_{m}}
\label{velocity of  center of mass $C_m$ of the robotic fish1}
\end{eqnarray}
\begin{eqnarray}
\frac{d\bmde{V_{C_m}}}{dt}=\frac{d\bmde{V_b}}{dt}\!+\!\frac{d\bmde{\omega_b}}{dt}\!\times\!\bmde{O_{b}C_{m}}\!+\!\bmde{\omega_b}\!\times\!\bmde{V_b}\!+\!\bmde{\omega_b}\!\times\!(\bmde{\omega_b}\!\times\!\bmde{O_{b}C_{m}})
\label{velocity of  center of mass $C_m$ of the robotic fish2}
\end{eqnarray}

Basing on Euler's equation, the total torque $\bmde{M_{total}}$ about $C_m$ is expressed as
\begin{eqnarray}
\left\{\begin{matrix}
\bmde{M_{total}}=\frac{d\bmde{H_{C_m}}}{dt}\\
\bmde{M_{total}}=\bmde{M_g}+\bmde{M_\omega}+\bmde{M_b^t}+\bmde{M_I}-\bmde{O_{b}C_{m}}\times\bmde{F_{total}}
\end{matrix}\right.
\label{the total torque acting on the robotic fish}
\end{eqnarray}
where $\bmde{H_{C_m}}$ is the moment of momentum about ${C_m}$ of the robotic fish, taking the form as
\begin{eqnarray}
\bmde{H_{C_m}}=\bmde{J}\bmde{\omega_{b}}+\bmde{M}\cdot\bmde{O_{b}C_{m}}\times\bmde{V_b}
\label{moment of momentum of the robotic fish1}
\end{eqnarray}
\begin{eqnarray}
\frac{d\bmde{H_{C_m}}}{dt}&=&\bmde{J}\dot{\bmde{\omega_{b}}}+\bmde{\omega_{b}}\times(\bmde{J}\bmde{\omega_b})+\bmde{M}\cdot(\bmde{\omega_b}\times\bmde{O_{b}C_{m}})\times \bmde{V_b}\nonumber\\&+&\bmde{M}\cdot\bmde{O_{b}C_{m}}\times(\dot{\bmde{V_b}}+\bmde{\omega_b}\times \bmde{V_b})
\label{moment of momentum of the robotic fish2}
\end{eqnarray}
$\bmde{J}=diag\left \{ {J}_{{xx}}, {J}_{{yy}}, {J}_{{zz}} \right \}$ is the moment of inertia about $O_{b}$ for the robotic fish, taking the form as
\begin{eqnarray}
\bmde{J}=\bmde{J_{ew}}+\bmde{J_{w}}
\label{moment of inertia of the robotic fish}
\end{eqnarray}
$\bmde{J_{w}}$ and $\bmde{J_{ew}}$ are the moments of inertia about $O_{b}$ for the weight block and the part apart from weight block, respectively, taking the form as
\begin{eqnarray}
\bmde{J_{\gamma}}=\bmde{{J_{\gamma}}'}&+&m_{\gamma}*diag\left \{ r_{O_{b}C_{\gamma}x}^2, r_{O_{b}C_{\gamma}y}^2, r_{O_{b}C_{\gamma}z}^2 \right \}
\label{moment of inertia of the robotic fish}
\end{eqnarray}
where $\gamma=ew, w$. '$ew$' and '$w$' indicate the part apart from the weight block and the weight block, respectively. $m_{ew}$ is mass of the part apart from the weight block, and $m_{ew}=m_{total}-m_{w}$. $r_{O_{b}C_{\gamma}x}$, $r_{O_{b}C_{\gamma}y}$, and $r_{O_{b}C_{\gamma}x}$ are components of the distance between $C_m$ and $C_{\gamma}$ along the $O_bx_b$ axis, $O_by_b$ axis, and $O_bz_b$ axis, respectively, taking the form as
\begin{eqnarray}
\begin {aligned}
r_{O_{b}C_{\gamma}x}^2=y_{C_{\gamma}}^2+z_{C_{\gamma}}^2\\
r_{O_{b}C_{\gamma}y}^2=x_{C_{\gamma}}^2+z_{C_{\gamma}}^2\\
r_{O_{b}C_{\gamma}z}^2=x_{C_{\gamma}}^2+y_{C_{\gamma}}^2
\end {aligned}
\end{eqnarray}
where $[x_{C_{ew}}, y_{C_{ew}}, z_{C_{ew}}]$ is coordinate of center of mass for the part apart from the weight block. $j_{C_{ew}} (j=x, y, z)$ takes the form as
\begin{eqnarray}
j_{C_{ew}}=M_{ew_{j}}/m_{ew}
\end{eqnarray}
$\bmde{{J_{\gamma}}'}$ is the moment of inertia about $C_{\gamma}$ for the part apart from weight block, taking the form as
\begin{eqnarray}
\bmde{{J_{\gamma}}'}=diag\left \{ {J}'_{\gamma_{xx}}, {J}'_{\gamma_{yy}}, {J}'_{\gamma_{zz}} \right \}
\end{eqnarray}

\begin{figure}[htb]
\centering
\includegraphics[width=0.9\columnwidth]{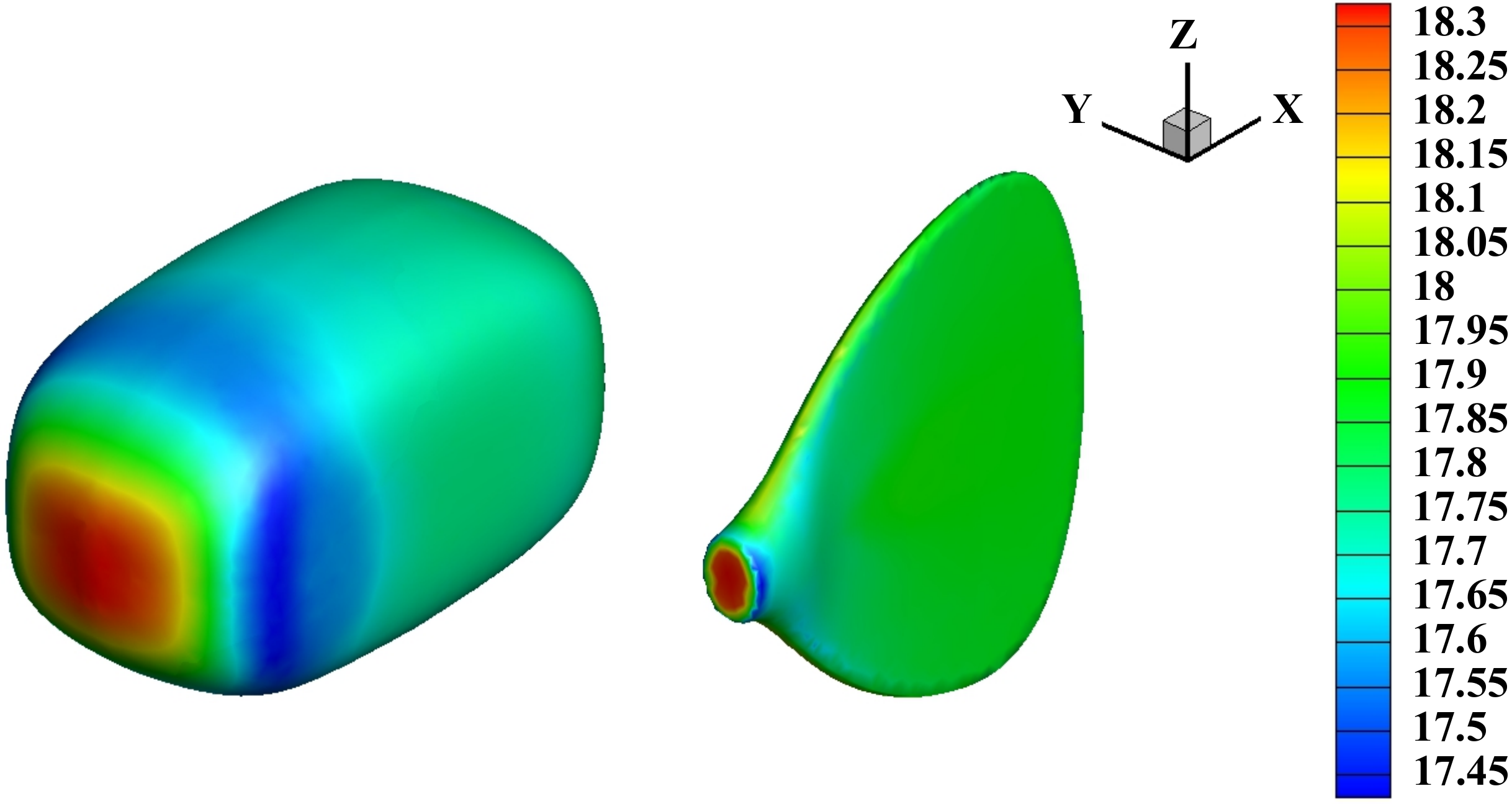}
\caption{\revisiontwo{Hydrodynamic pressure variations on the surface of the fish body and tail when $\alpha _{b_1}$, $\alpha _{b_2}$, and $\alpha _{t}$ are 0.}}
\label{Dynamic pressure of the robotic fish}
\end{figure}
\revisiontwo{Basing on the above analyses, the concrete form of the dynamic equations (\ref{the total force acting on the robotic fish}) and (\ref{the total torque acting on the robotic fish}) can be finally acquired, as shown in (\ref{the total forces and torques about the $O_bx_b$ axis, the $O_by_b$ axis, and the $O_bz_b$ axis}) where $F_{x_b}$, $F_{y_b}$, $F_{z_b}$ are components of the total force along the $O_bx_b$ axis, $O_by_b$ axis, and $O_bz_b$ axis, respectively. $M_{x_b}$, $M_{y_b}$, $M_{z_b}$ are components of the total torque about the $O_bx_b$ axis, $O_by_b$ axis, and $O_bz_b$ axis, respectively.}
\newcounter{mytempeqncnt}
\begin{figure*}[tb]
\normalsize
\setcounter{mytempeqncnt}{\value{equation}}
\setcounter{equation}{45}
\begin{small}
\begin{eqnarray}
\revisiontwo{\left\{\begin{matrix}
F_{x_b}=m_{total}\cdot\left [ \dot{V_{b_x}}-V_{b_y}\cdot \omega_{b_z}+V_{b_z}\cdot \omega_{b_y}-x_{C_m}\left ( \omega_{b_z}^2+\omega_{b_y}^2 \right )+y_{C_m}\left ( \omega_{b_x}\omega_{b_y}-\dot{\omega_{b_z}} \right )+z_{C_m}\left ( \omega_{b_x}\omega_{b_z}+\dot{\omega_{b_y}} \right ) \right ]\\
F_{y_b}=m_{total}\cdot\left [ \dot{V_{b_y}}-V_{b_z}\cdot \omega_{b_x}+V_{b_x}\cdot \omega_{b_z}-y_{C_m}\left ( \omega_{b_x}^2+\omega_{b_z}^2 \right )+z_{C_m}\left ( \omega_{b_y}\omega_{b_z}-\dot{\omega_{b_x}} \right )+x_{C_m}\left ( \omega_{b_x}\omega_{b_y}+\dot{\omega_{b_z}} \right ) \right ]\\
F_{z_b}=m_{total}\cdot\left [ \dot{V_{b_z}}-V_{b_x}\cdot \omega_{b_y}+V_{b_y}\cdot \omega_{b_x}-z_{C_m}\left ( \omega_{b_y}^2+\omega_{b_x}^2 \right )+x_{C_m}\left ( \omega_{b_z}\omega_{b_x}-\dot{\omega_{b_y}} \right )+y_{C_m}\left ( \omega_{b_y}\omega_{b_z}+\dot{\omega_{b_x}} \right ) \right ]\\
M_{x_b}=J_{xx}\dot{\omega_{b_x}}+\left ( J_{zz}-J_{yy} \right )\omega_{b_y}\omega_{b_z}+m_{total}\cdot\left [ y_{C_m}\left ( \dot{V_{b_z}}+V_{b_y}\omega_{b_x}-V_{b_x}\omega_{b_y} \right )-z_{C_m}\left ( \dot{V_{b_y}}+V_{b_x}\omega_{b_z}-V_{b_z}\omega_{b_x} \right ) \right ]\\
M_{y_b}=J_{yy}\dot{\omega_{b_y}}+\left ( J_{xx}-J_{zz} \right )\omega_{b_z}\omega_{b_x}+m_{total}\cdot\left [ z_{C_m}\left ( \dot{V_{b_x}}+V_{b_z}\omega_{b_y}-V_{b_y}\omega_{b_z} \right )-x_{C_m}\left ( \dot{V_{b_z}}+V_{b_y}\omega_{b_x}-V_{b_x}\omega_{b_y} \right ) \right ]\\
M_{z_b}=J_{zz}\dot{\omega_{b_z}}+\left ( J_{yy}-J_{xx} \right )\omega_{b_x}\omega_{b_y}+m_{total}\cdot\left [ x_{C_m}\left ( \dot{V_{b_y}}+V_{b_x}\omega_{b_z}-V_{b_z}\omega_{b_x} \right )-y_{C_m}\left ( \dot{V_{b_x}}+V_{b_z}\omega_{b_y}-V_{b_y}\omega_{b_z} \right ) \right ]
\end{matrix}\right.}
\label{the total forces and torques about the $O_bx_b$ axis, the $O_by_b$ axis, and the $O_bz_b$ axis}
\end{eqnarray}
\end{small}
\setcounter{equation}{\value{mytempeqncnt}}
\vspace*{1pt} 
\end{figure*}
\begin{figure}[htb]
\centering
\begin{tabular}{cc}
\begin{minipage}{0.48\linewidth}
\centering
\includegraphics[width=\columnwidth]{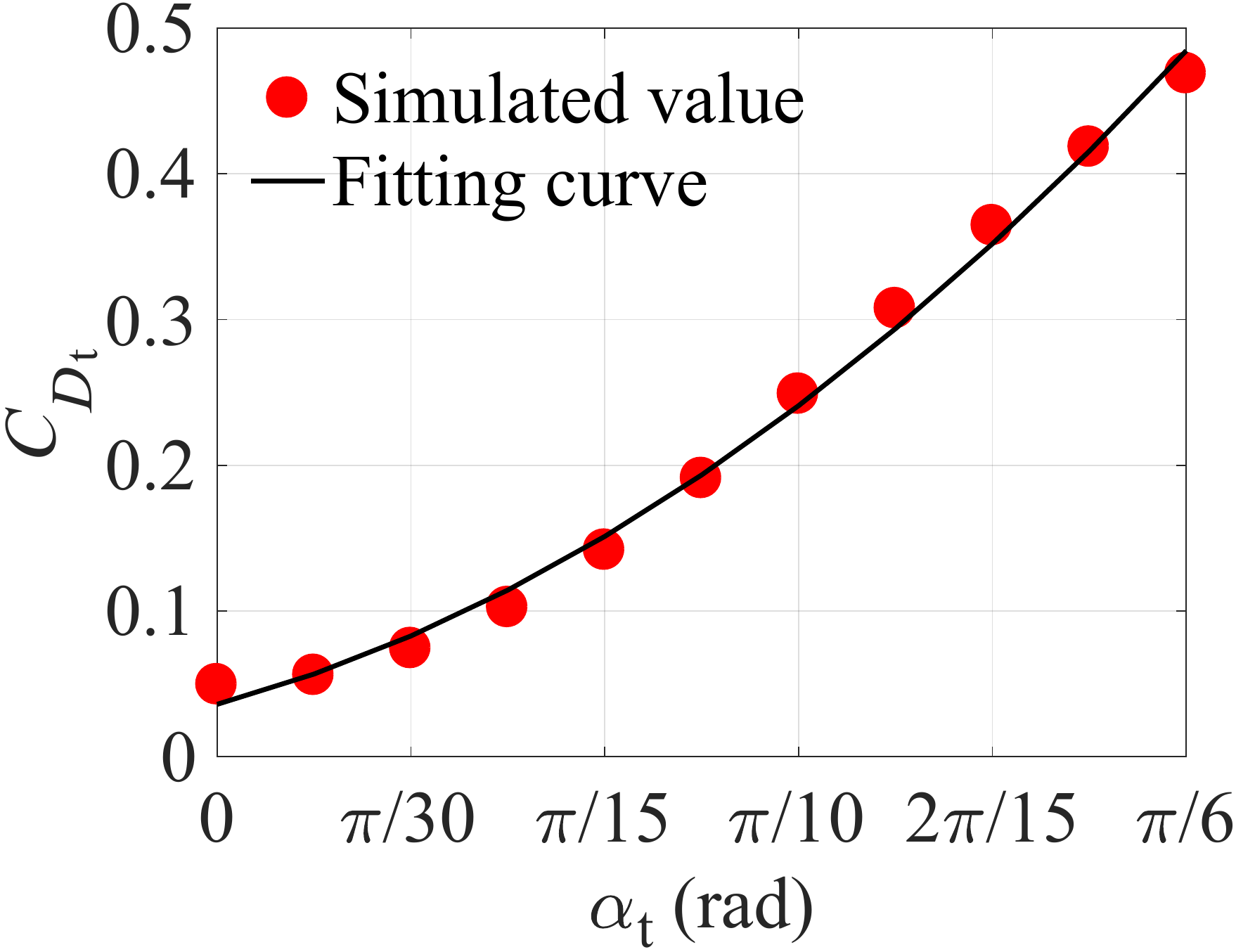}\\
(a)
\end{minipage}
\begin{minipage}{0.48\linewidth}
\centering
\includegraphics[width=\columnwidth]{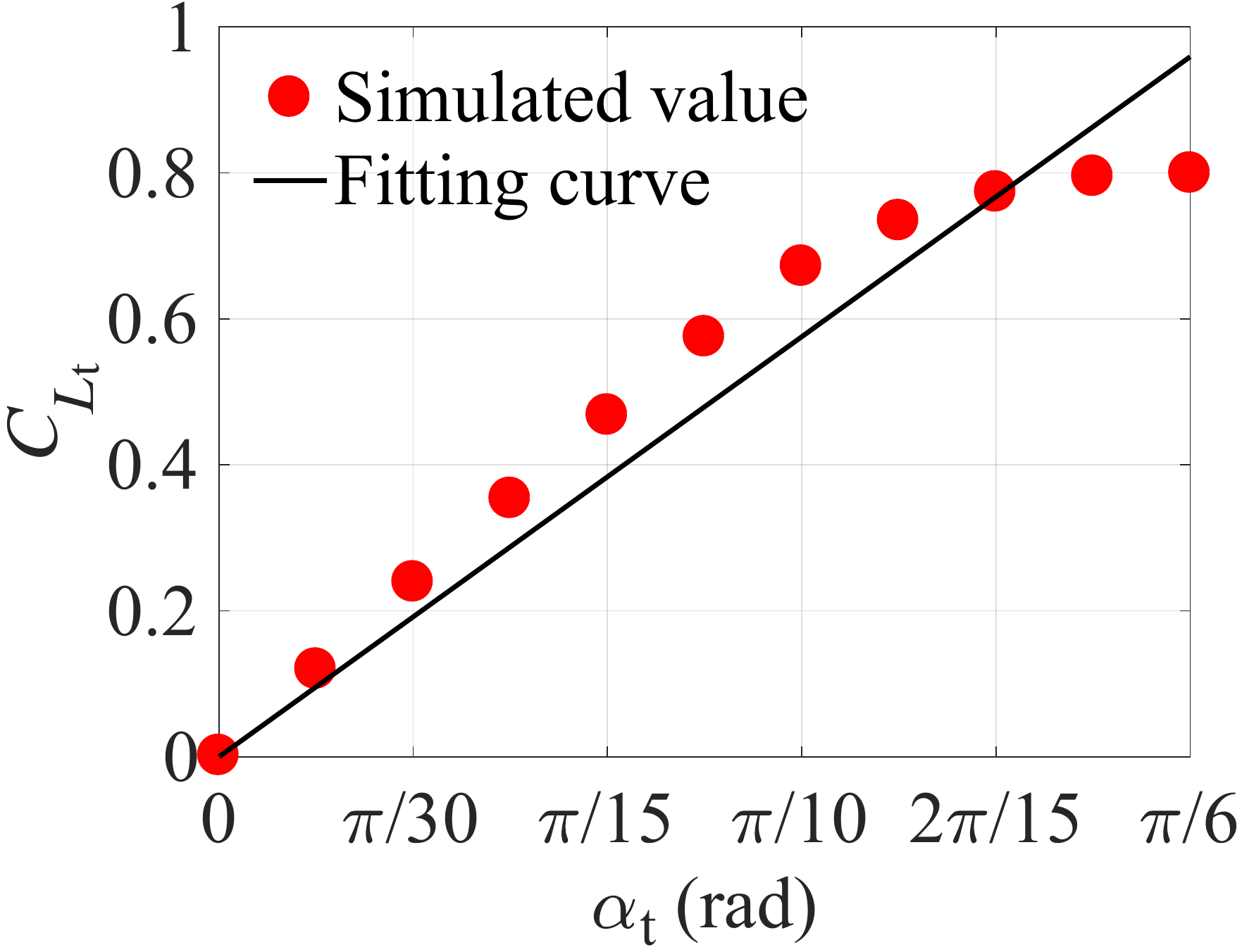}\\
(b)
\end{minipage}
\end{tabular}\\
\begin{tabular}{cc}
\begin{minipage}{0.48\linewidth}
\centering
\includegraphics[width=\columnwidth]{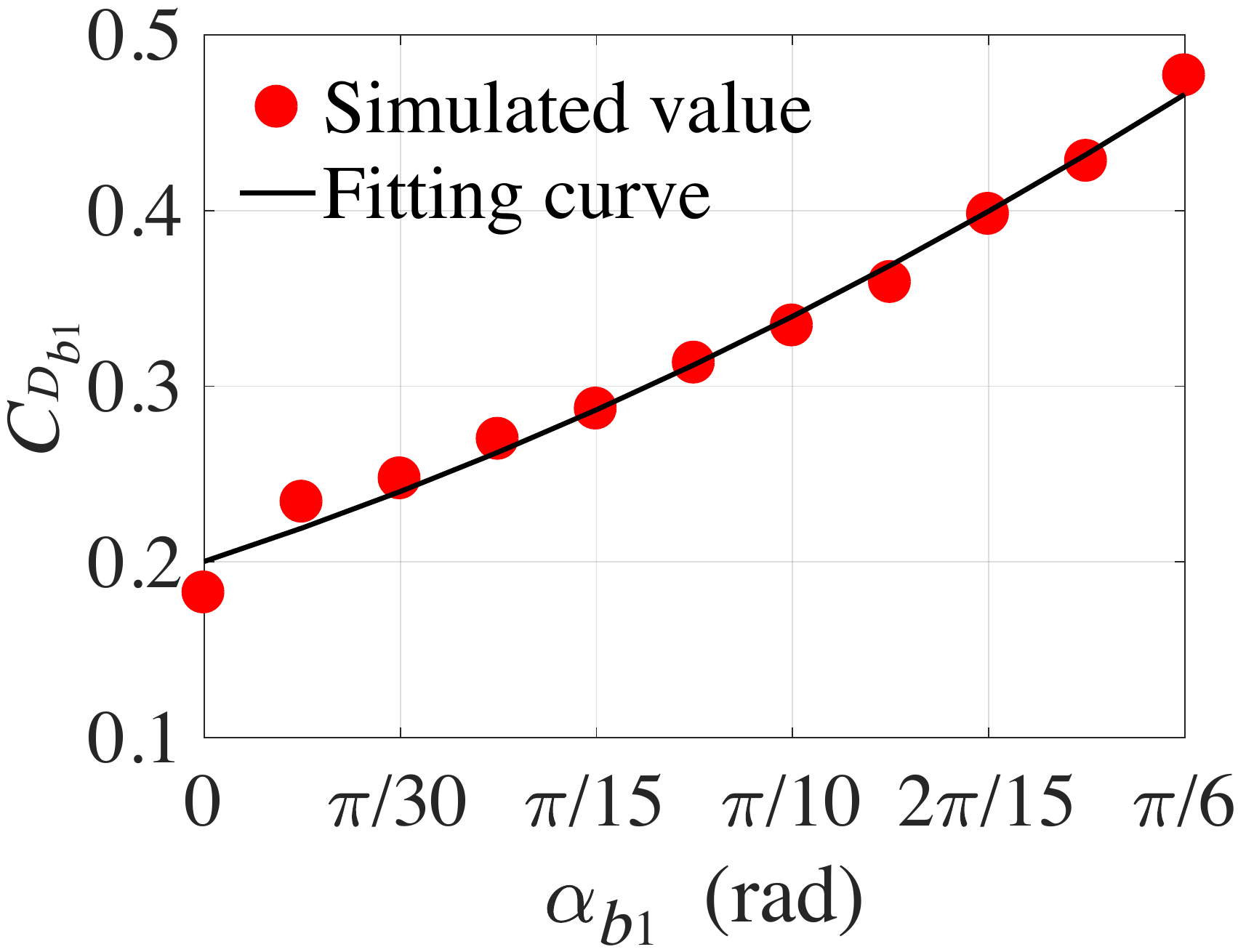}\\
(c)
\end{minipage}
\begin{minipage}{0.48\linewidth}
\centering
\includegraphics[width=\columnwidth]{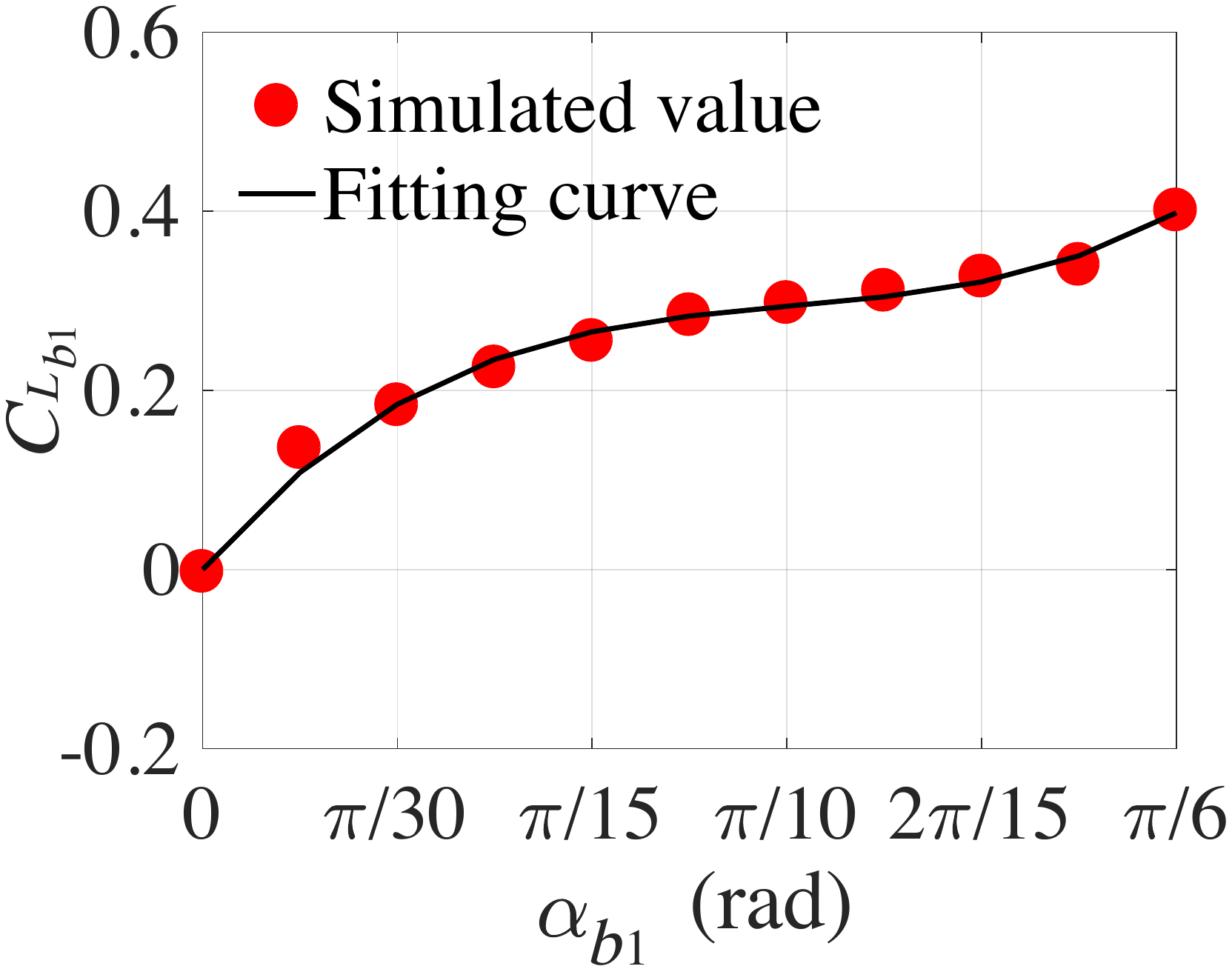}\\
(d)
\end{minipage}
\end{tabular}\\
\begin{tabular}{cc}
\begin{minipage}{0.48\linewidth}
\centering
\includegraphics[width=\columnwidth]{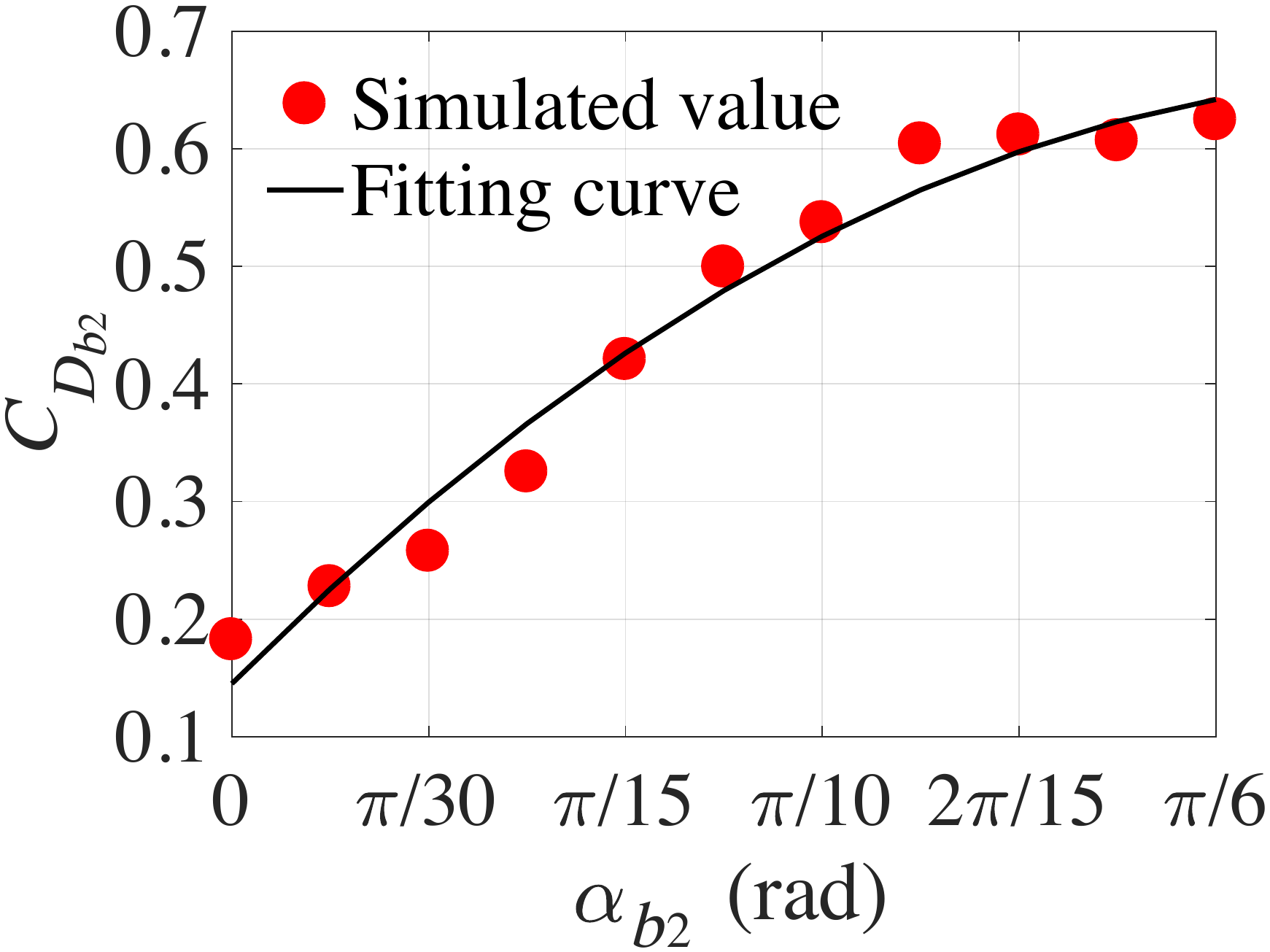}\\
(e)
\end{minipage}
\begin{minipage}{0.48\linewidth}
\centering
\includegraphics[width=\columnwidth]{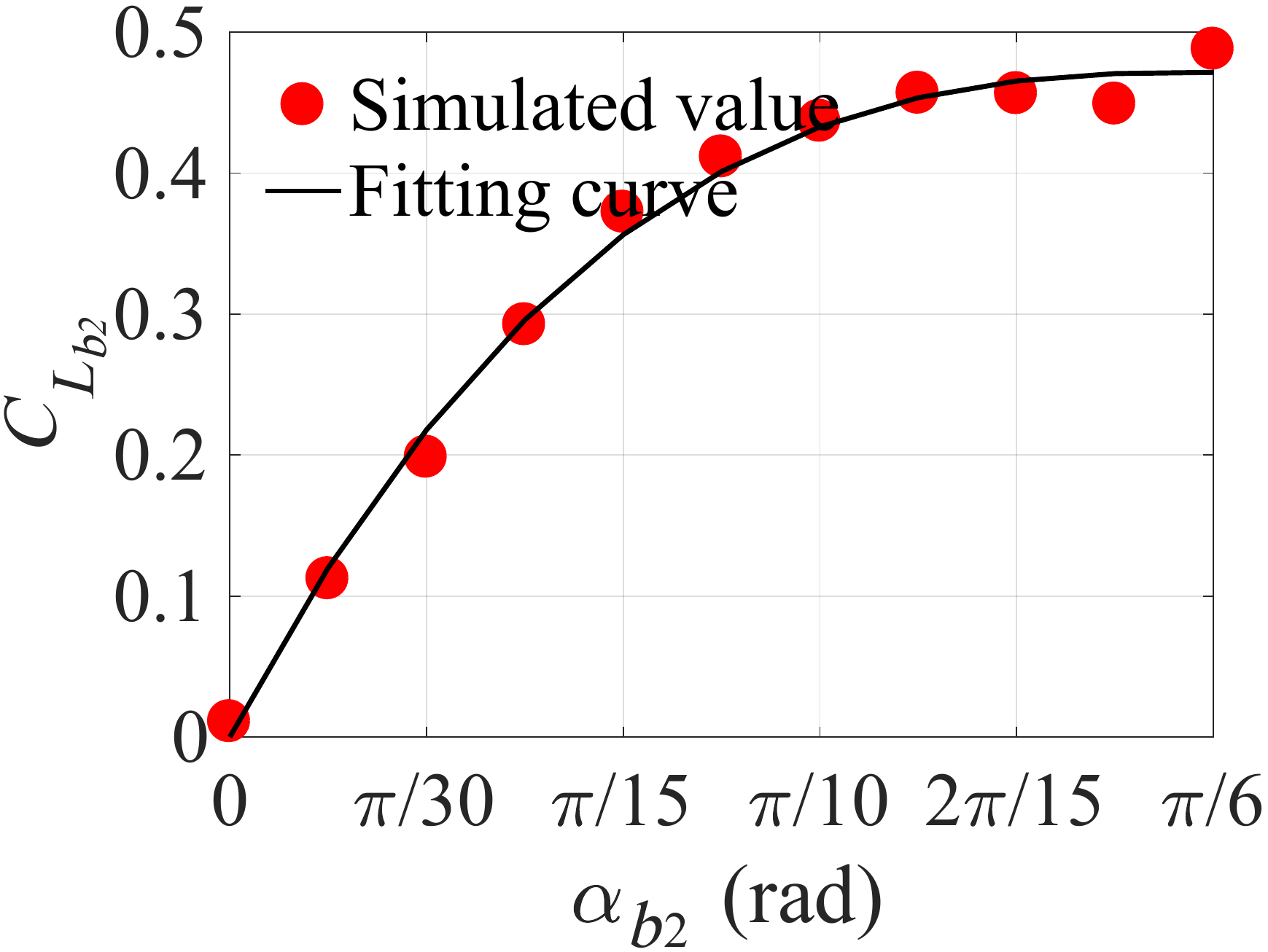}\\
(f)
\end{minipage}
\end{tabular}\\
\begin{tabular}{cc}
\begin{minipage}{0.48\linewidth}
\centering
\includegraphics[width=\columnwidth]{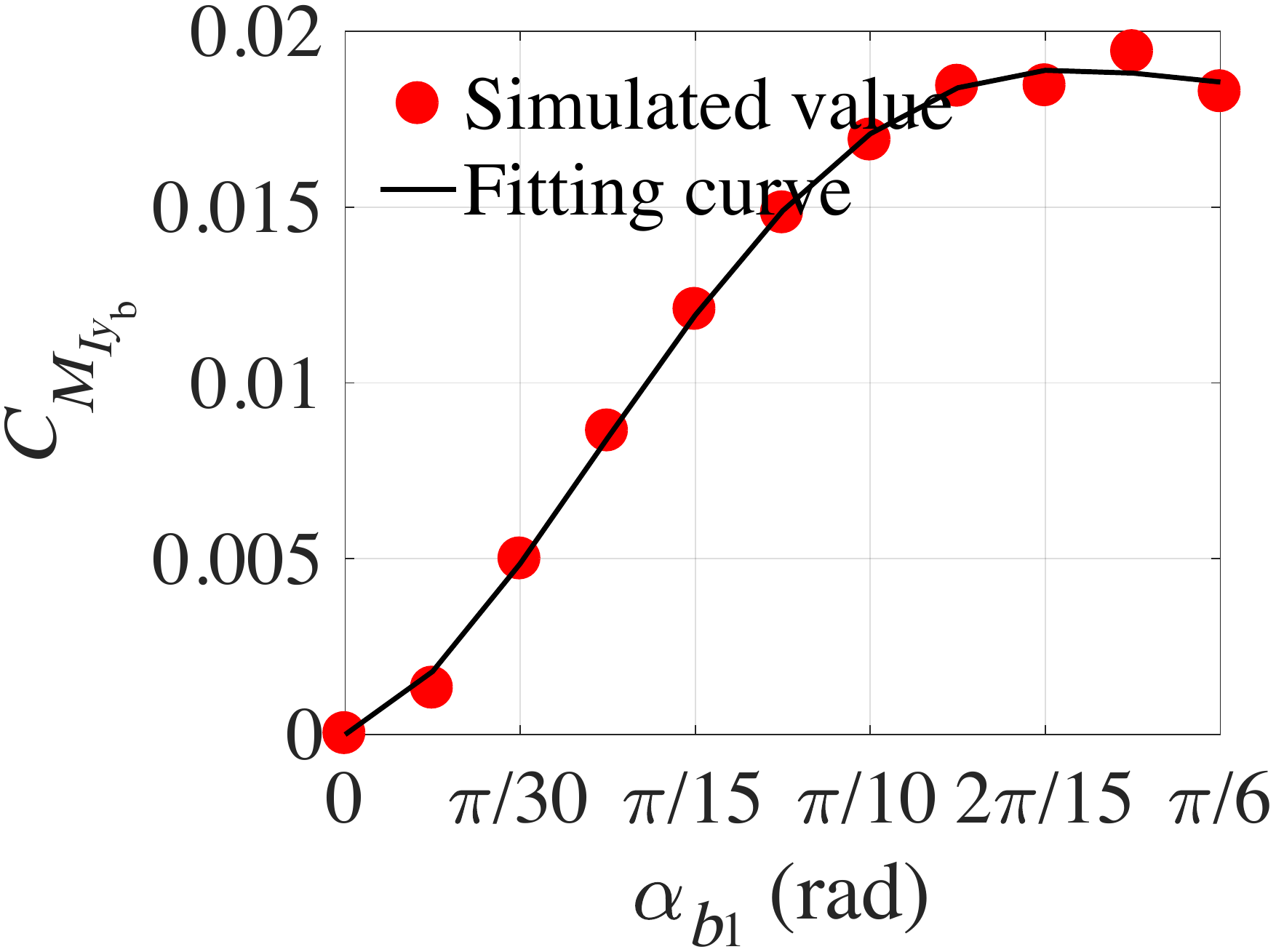}\\
(g)
\end{minipage}
\begin{minipage}{0.48\linewidth}
\centering
\includegraphics[width=\columnwidth]{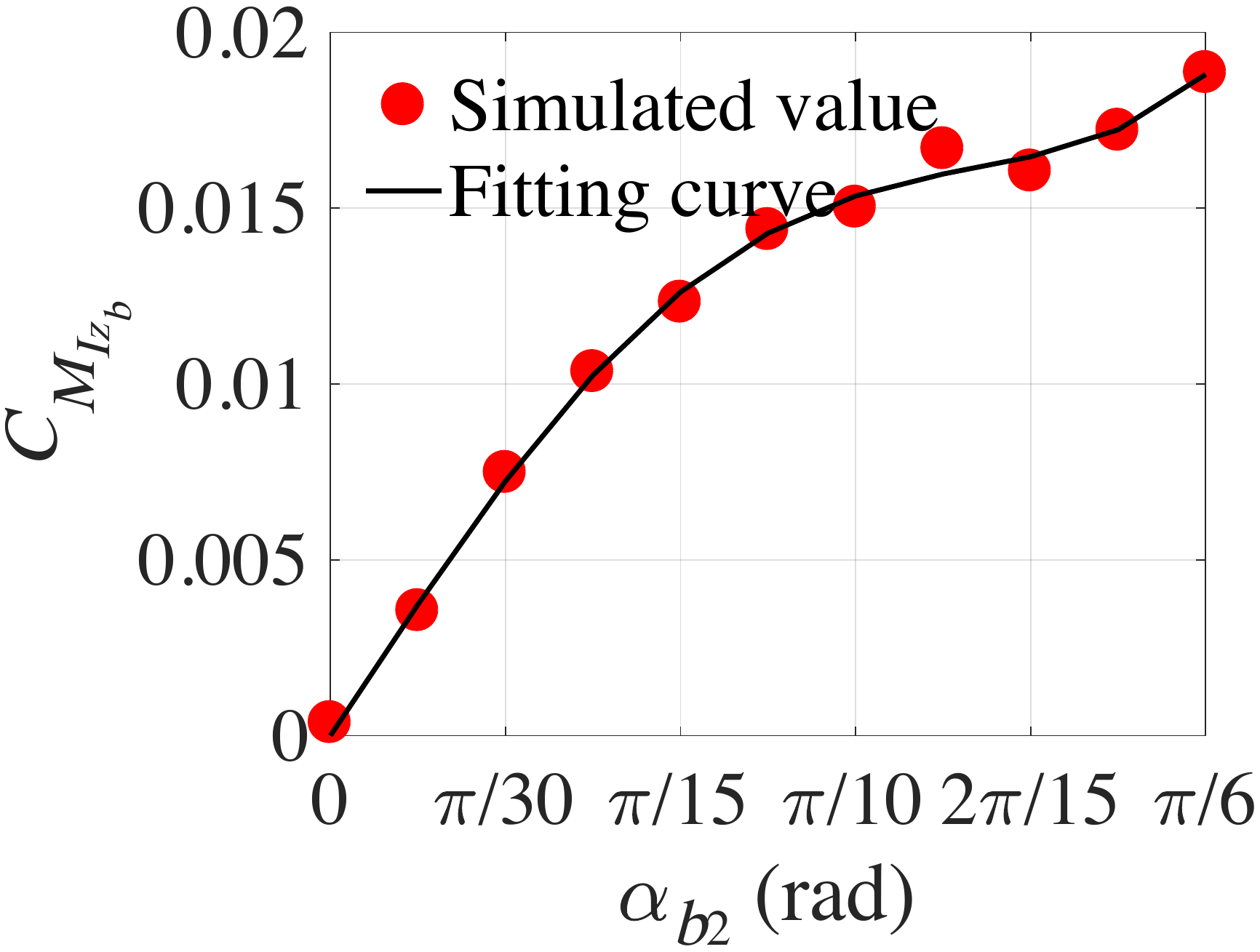}\\
(h)
\end{minipage}
\end{tabular}
\caption{The lift, drag, and impact torque coefficients acquired by computational fluid dynamics (CFD) simulation. (a) $C_{D_{t}}$. (b) $C_{L_{t}}$. (c) $C_{D_{b_1}}$. (d)  $C_{L_{b_1}}$. (e) $C_{D_{b_2}}$. (f) $C_{L_{b_2}}$. (g) $C_{M_{I_{y_b}}}$. (h) $C_{M_{I_{z_b}}}$.}
\label{Determination of Force Coefficients and Torque Coefficients}
\end{figure}
\subsection{Determination of Model Parameters}
In this part, model parameters\revisiontwo{,} which include mass, dimensions, and moment of inertia of the robotic fish, etc. are determined by three-dimensional computer-aided design (CAD) software SolidWorks, as shown in \revisiontwo{Table S1 of the supplementary materials}. We have used two robotic fish to conduct the experiments. $m_{b_1}$ is buoyancy mass for the robotic fish used in rectilinear motion and turning motion, while $m_{b_2}$ is for the robotic fish used in gliding motion and spiral motion. $m_{b_1}$ and $m_{b_2}$ are both determined by actual measurement. Lift coefficients, drag coefficients, and impact torque coefficients are determined by computational fluid dynamics (CFD) simulation. Damping torque coefficients are determined by grey-box model estimation method.
\subsubsection{Determining Force Coefficients and Torque Coefficients Using Computational Fluid Dynamics (CFD) Method}
Specifically, \revision{computational fluid dynamics} (CFD) simulation for fish body and tail of the robotic fish were respectively conducted using a CFD software called HyperFlow, which is developed by China Aerodynamics Research and Development Center (CARDC). HyperFlow is a structured/unstructured hybrid integrated fluid simulation software. It is able to \revisiontwo{run the structured solver synchronously} on structured grids and unstructured solver on unstructured grids. \revisiontwo{Besides,} it has been proved to have good performance in multi-purpose fluid simulation \cite{he2015hyperflow,he2016validation}.
Figure~\ref{Dynamic pressure of the robotic fish} shows the hydrodynamic pressure variations of the tail and fish body using CFD simulation. More details about the CFD simulation can be found in \revisiontwo{Section S1 of the supplementary materials}. In the \revisiontwo{CFD simulation}, angles of attack $\alpha_{t}$, $\alpha_{b_1}$, and $\alpha_{b_2}$ changed from 0 to $\pi/6$ rad with an interval of $\pi/60$ rad. Basing on the hydrodynamic pressure variations, the lift, drag, and impact torque coefficients under certain values of $\alpha_{t}$, $\alpha_{b_1}$, and $\alpha_{b_2}$ are acquired, as shown in Figure~\ref{Determination of Force Coefficients and Torque Coefficients}. Basing on data fitting method, the quantitative equations which link $\alpha_{t}$/$\alpha_{b_1}$/$\alpha_{b_2}$ to \revisiontwo{coefficients mentioned above} can be acquired, as shown in Section S1 of the supplementary materials.
\subsubsection{Determining the damping torque coefficients using grey-box model estimation method}
The damping torque coefficients are determined by grey-box model estimation method \cite{ljung1995system}. In the grey-box model estimation, we recorded the rectilinear velocity of the robotic fish with given oscillating parameters\revisiontwo{,} including amplitude and frequency of the tail in 28 s. \revisiontwo{The} input data for grey-box model were the oscillating parameters, while the output data were the rectilinear velocity. As shown in Table S2 of the supplementary materials, we restricted ranges of the three coefficients for avoiding drift of the solution. \revisiontwo{The} final values of the damping coefficients are shown in Table S2 of the supplementary materials. Figure~\ref{Measured velocity and simulated velocity obtained using the estimated coefficients} shows the measured velocity and simulated velocity obtained using the estimated coefficients. The measured velocity and simulated velocity of the robotic fish match with a 61.45\% fit.
\begin{figure}[htb]
\centering
\includegraphics[width=0.75\linewidth]{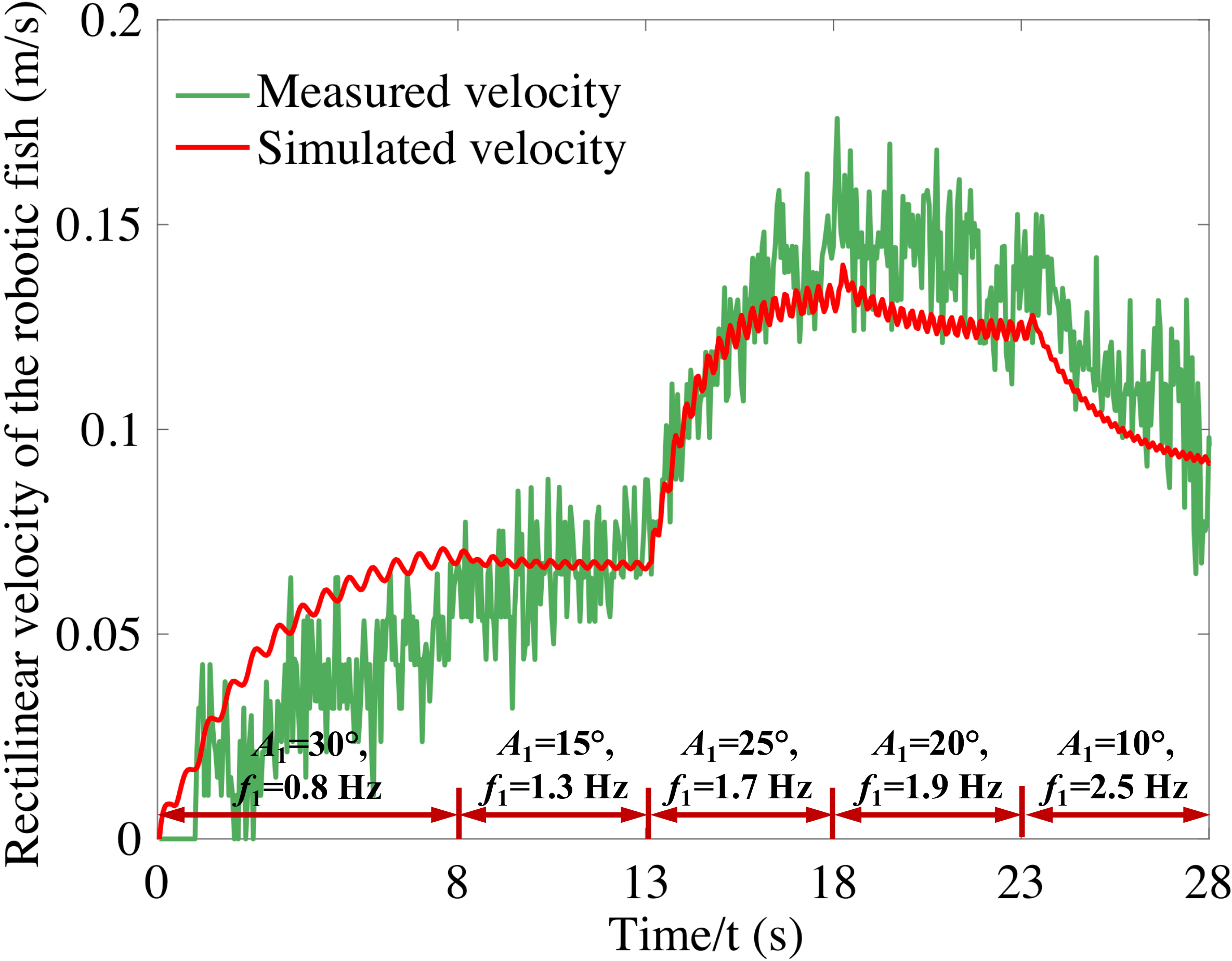}
\caption{Measured rectilinear velocity and estimated rectilinear velocity obtained using grey-box model estimation method.}
\label{Measured velocity and simulated velocity obtained using the estimated coefficients}
\end{figure}
\section{Simulations and experiments}
\subsection{Rectilinear motion}
\begin{figure}[htb]
\centering
\begin{tabular}{cc}
\begin{minipage}{0.495\linewidth}
\centering
\includegraphics[width=\columnwidth]{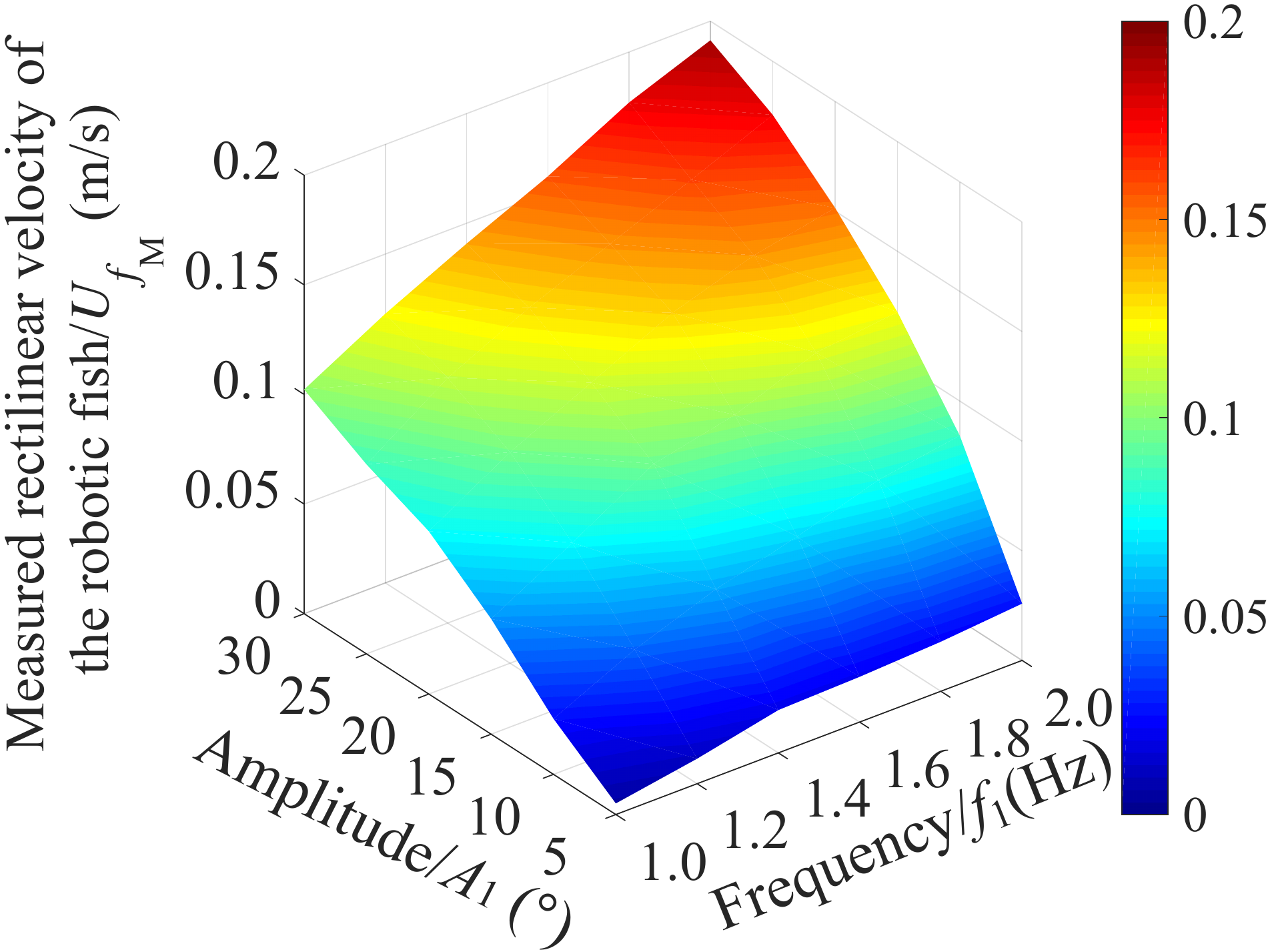}\\
(a)
\end{minipage}
\begin{minipage}{0.495\linewidth}
\centering
\includegraphics[width=\columnwidth]{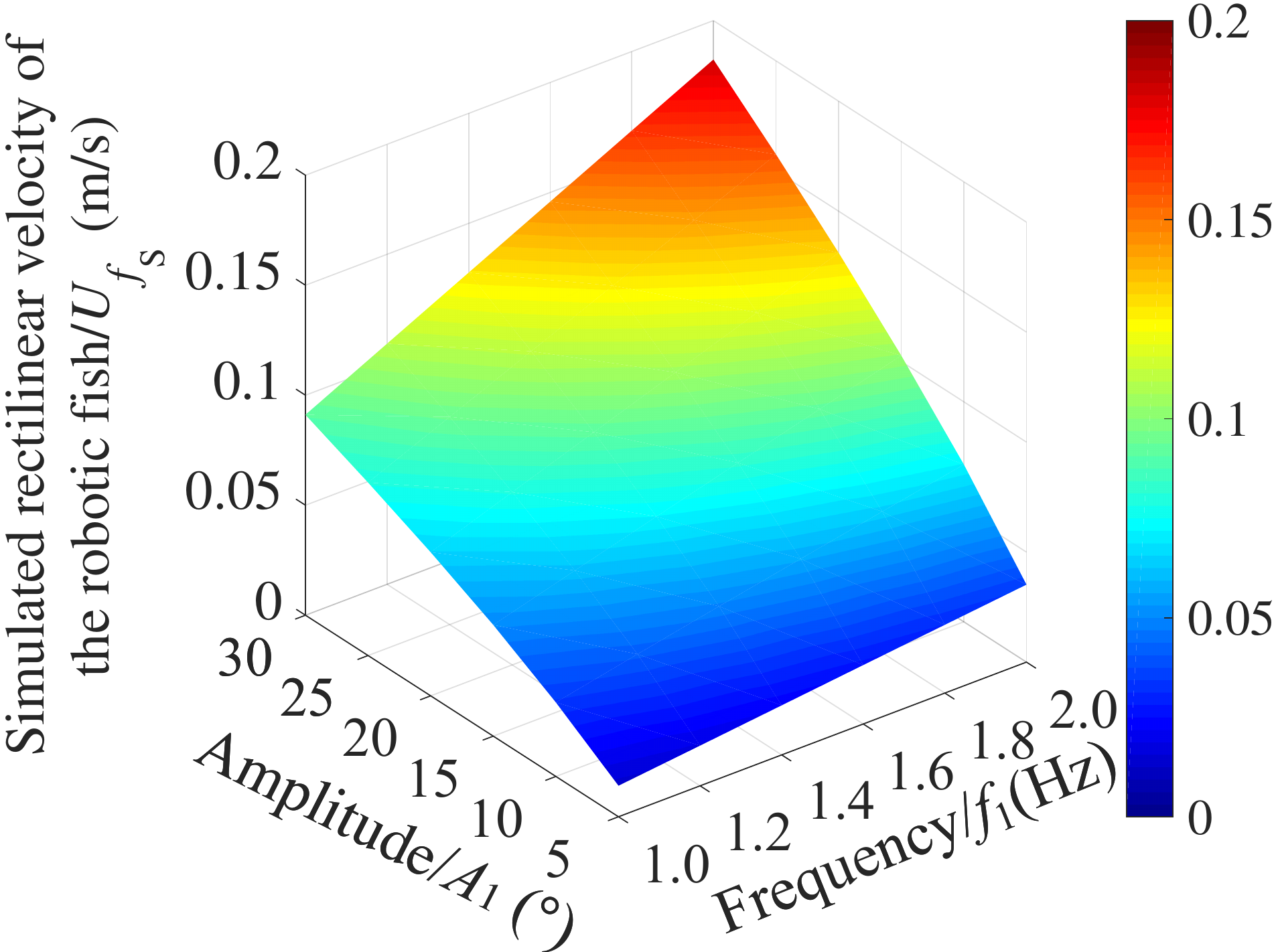}\\
(b)
\end{minipage}
\end{tabular}
\caption{Measured and simulated rectilinear motion velocity of the robotic fish. (a) Measured value. (b) Simulated value.}
\label{Measured and simulated rectilinear motion velocity of the robotic fish}
\end{figure}
\begin{figure}[htb]
\centering
\includegraphics[width=0.83\linewidth]{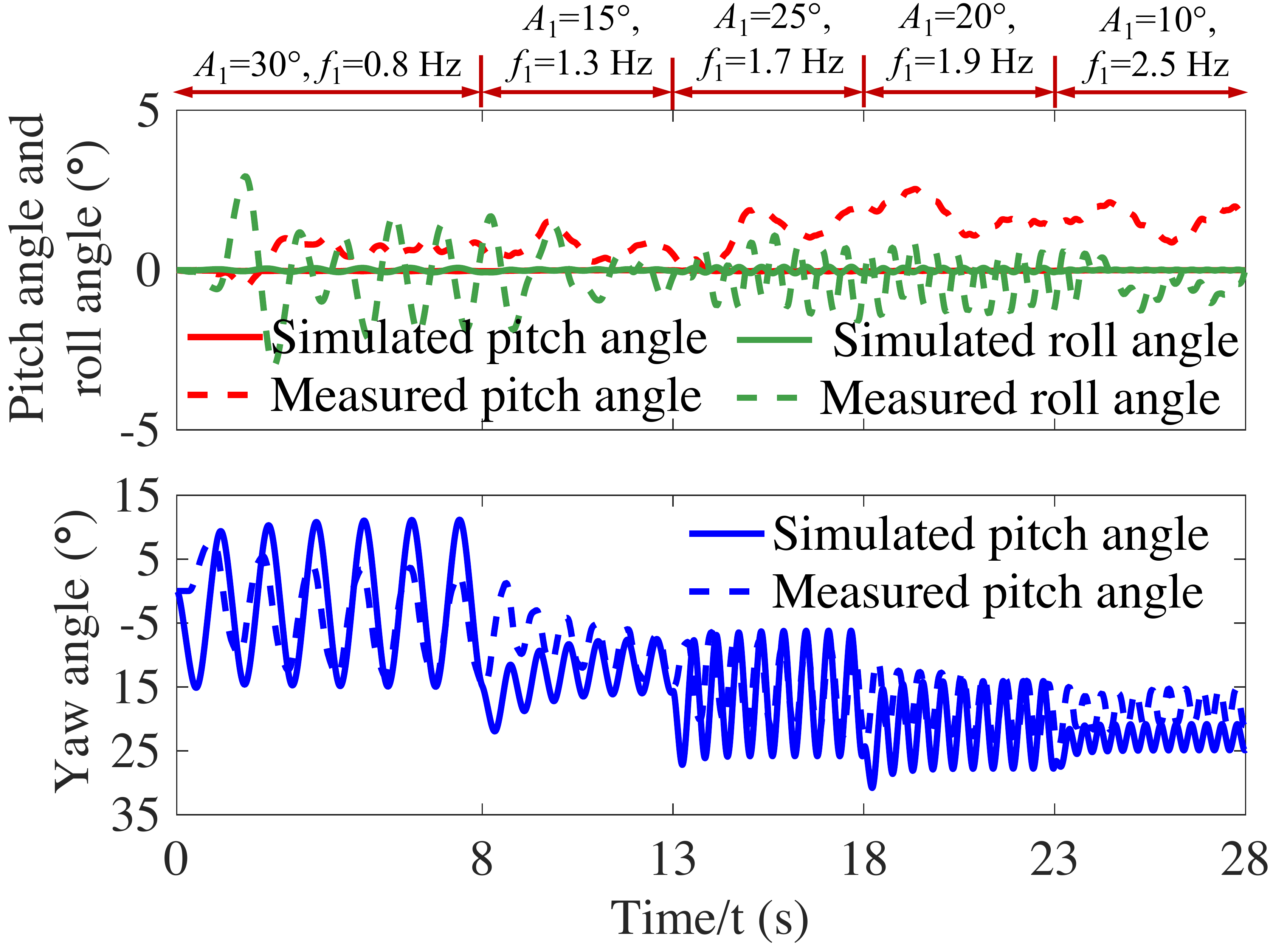}\\
\caption{\revisiontwo{Real-time attitudes of the robotic fish in rectilinear motion. }}
\label{Attitude angle of the robotic fish in rectilinear motion}
\end{figure}
\begin{figure}[htb]
\centering
\includegraphics[width=0.83\linewidth]{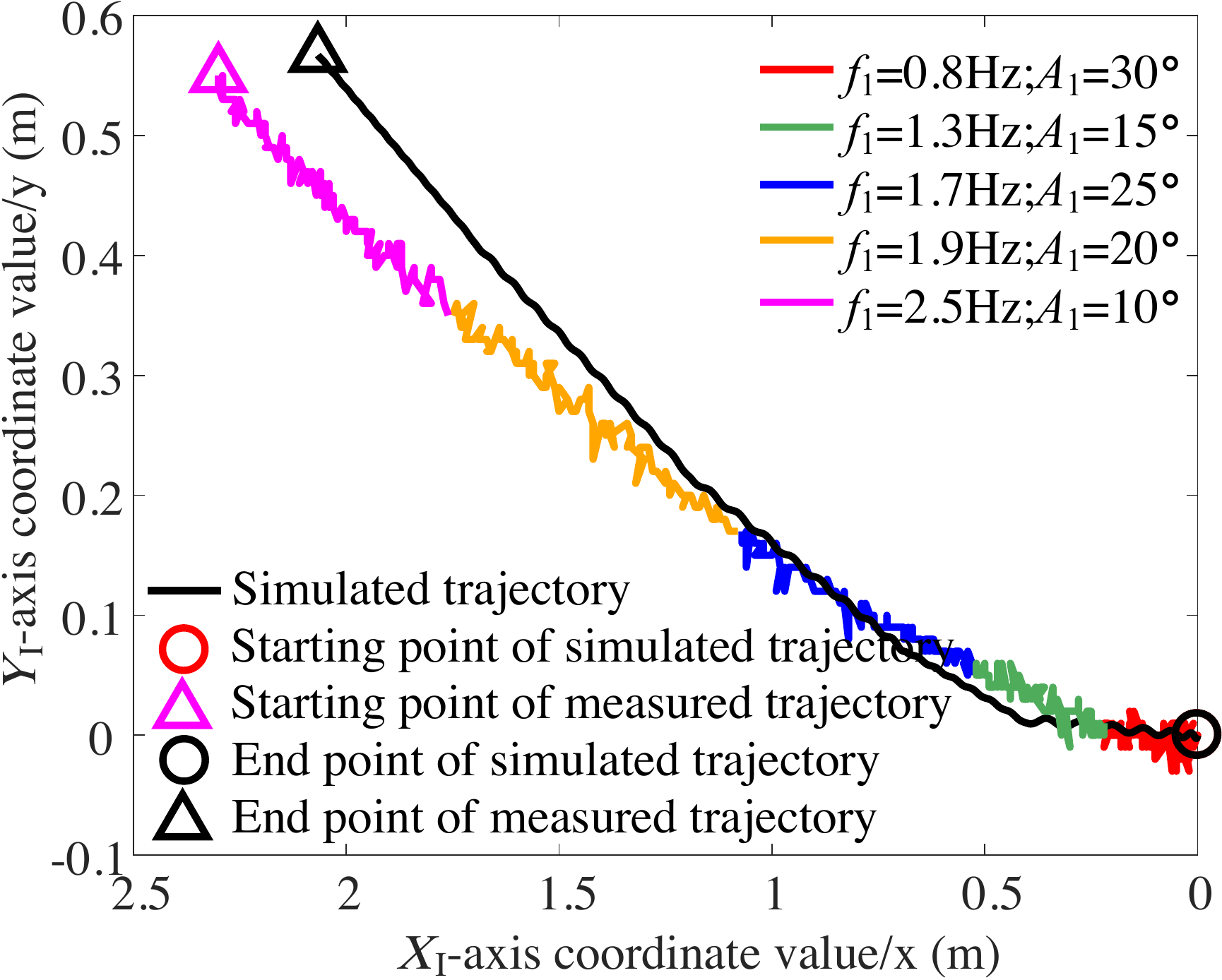}\\
\caption{Trajectory of the robotic fish under five combinations of $A_1$ and $f_1$.}
\label{Trajectory of the robotic fish in rectilinear motion}
\end{figure}
In rectilinear motion experiment, varieties of rectilinear velocities were obtained by changing the oscillating frequency $f_1$ and amplitude $A_1$ of the tail, while the oscillating offset $\bar{\xi_{1}}$ was zero. Figure~\ref{Measured and simulated rectilinear motion velocity of the robotic fish} shows the measured and simulated rectilinear motion velocity $U_r$ obtained by various combinations of $f_1$ and $A_1$. $U_r$ is the resultant velocity of the velocity $V_{I_x}$ along the axis $O_IX_I$ and the velocity $V_{I_y}$ along the axis $O_IY_I$. \revisiontwo{It} increases with $f_1$ and $A_1$. The measured and simulated $U_r$ match well with a coefficient of determination ($R^2$) of 0.8898 and a mean absolute error (MAE) of 0.0137 m/s. Figure~\ref{Attitude angle of the robotic fish in rectilinear motion} shows the real-time attitude of the robotic fish when it was actuated by five combinations of $A_1$ and $f_1$. Under each combination of $A_1$ and $f_1$, yaw angle of the robotic fish oscillates around a certain value while roll and pitch angle of the robotic fish oscillate around zero, \revision{in which case} the robotic fish swims in a straight line. Because of the periodical oscillation of the tail, the robotic fish body oscillates while swimming. Thus the yaw angle, pitch angle\revisiontwo{,} and roll angle of the robotic fish oscillate periodically with the time. It can be seen that the simulated and measured attitudes match well in the oscillatory feature and value. A more careful inspection revealed that the yaw amplitude increases with the increasing $A_1$ while the yaw rate increases with the increasing $f_1$. \revision{For pitch angle and roll angle, the biggest errors between the estimated values and the measured values are both less than 3$^\circ$, which are small enough. The errors are results of the wave motion of water  which caused the roll motion and pitch motion of the robotic fish.} \revisiontwo{The} final trajectory of the robotic fish is shown in Figure~\ref{Trajectory of the robotic fish in rectilinear motion}, with a maximum error between the simulated trajectory and measured trajectory of 0.2407 m.
\subsection{Turning motion}
\begin{figure}[htb]
\centering
\begin{tabular}{cc}
\begin{minipage}{0.495\linewidth}
\centering
\includegraphics[width=\columnwidth]{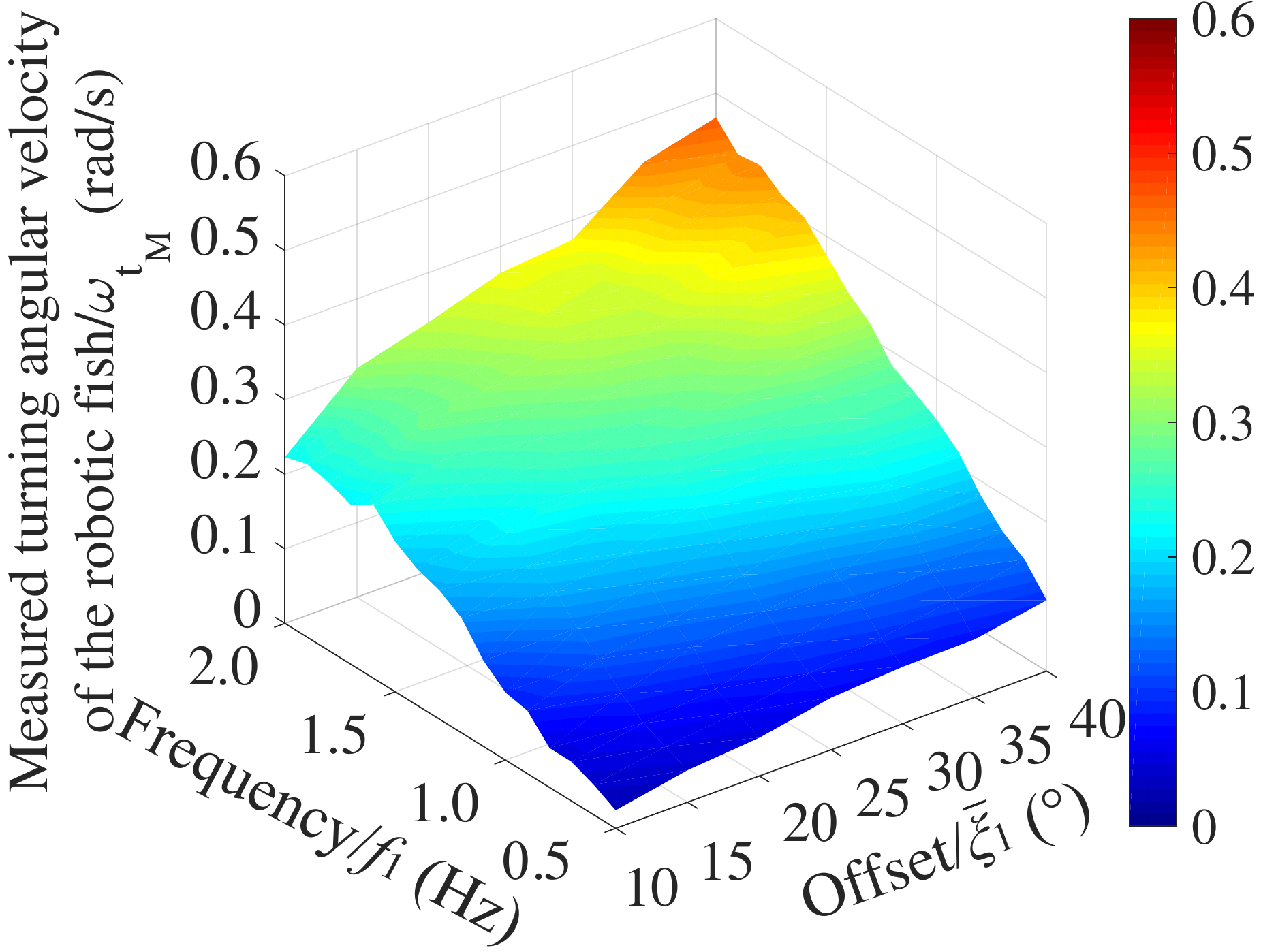}\\
(a)
\end{minipage}
\begin{minipage}{0.495\linewidth}
\centering
\includegraphics[width=\columnwidth]{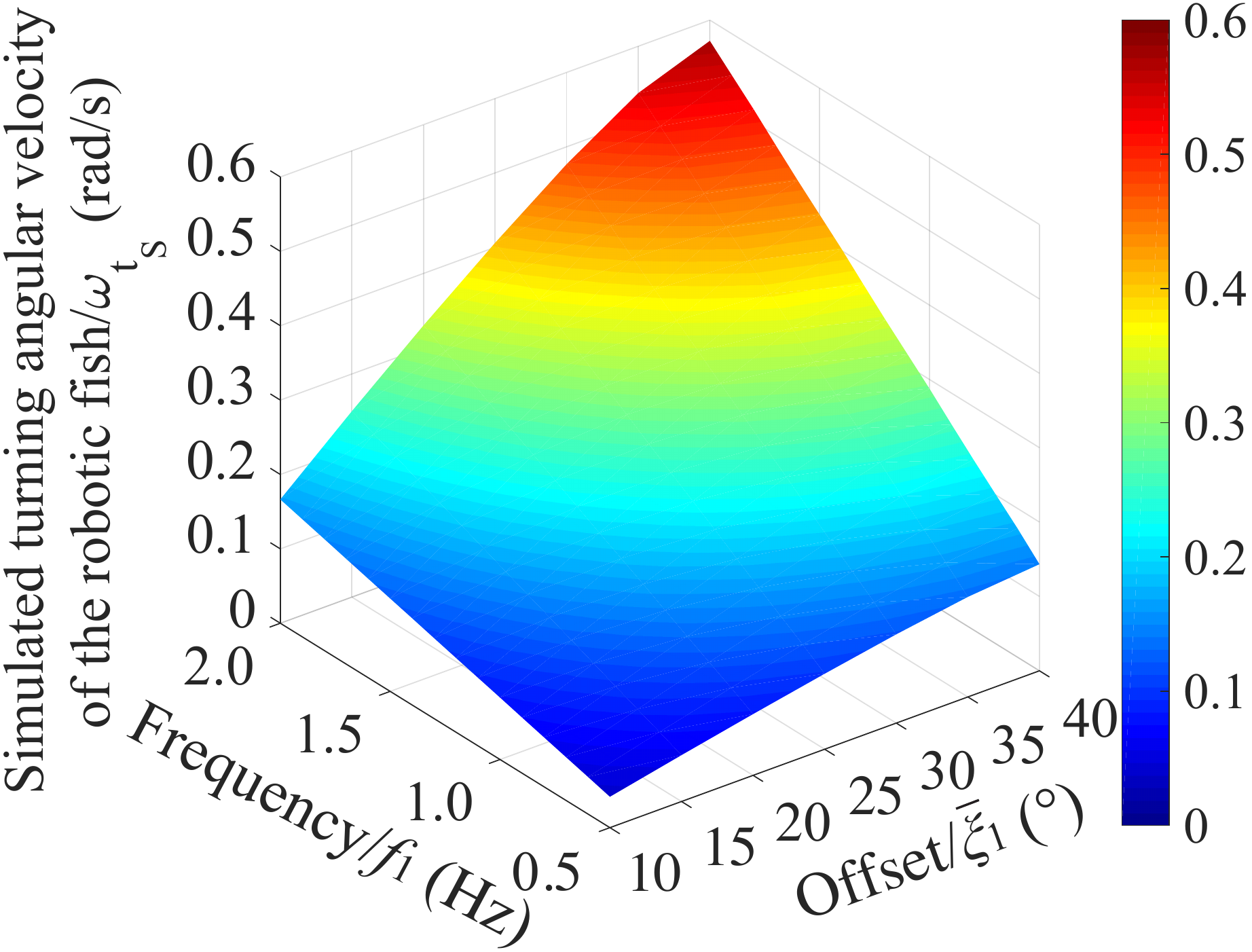}\\
(b)
\end{minipage}
\end{tabular}
\caption{Measured and simulated turning angular velocity of the robotic fish. (a) Measured value. (b) Simulated value.}
\label{Actual turning angular velocity of the robotic fish}
\end{figure}
\begin{figure}[htb]
\centering
\begin{tabular}{cc}
\begin{minipage}{0.495\linewidth}
\centering
\includegraphics[width=\columnwidth]{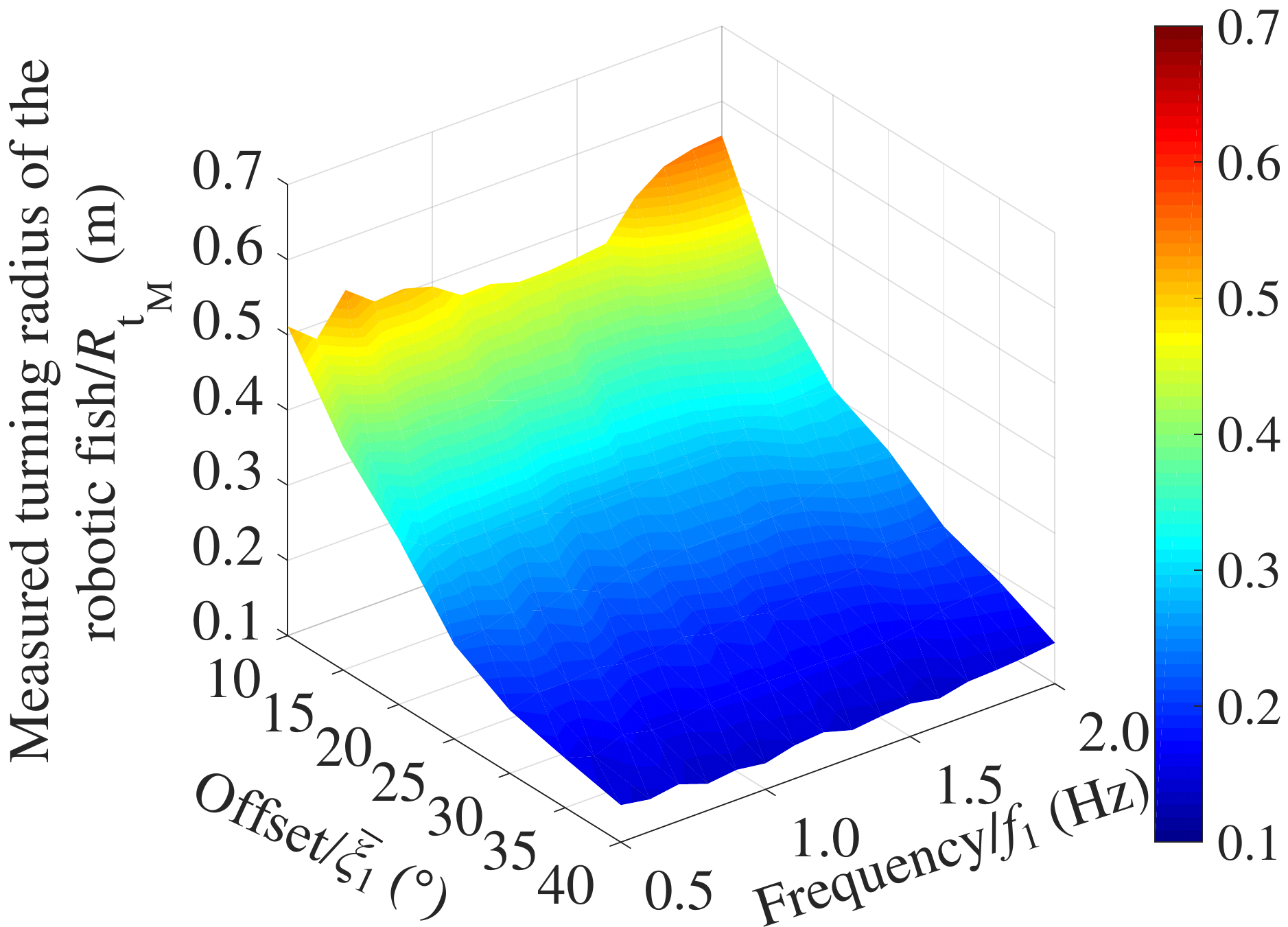}\\
(a)
\end{minipage}
\begin{minipage}{0.495\linewidth}
\centering
\includegraphics[width=\columnwidth]{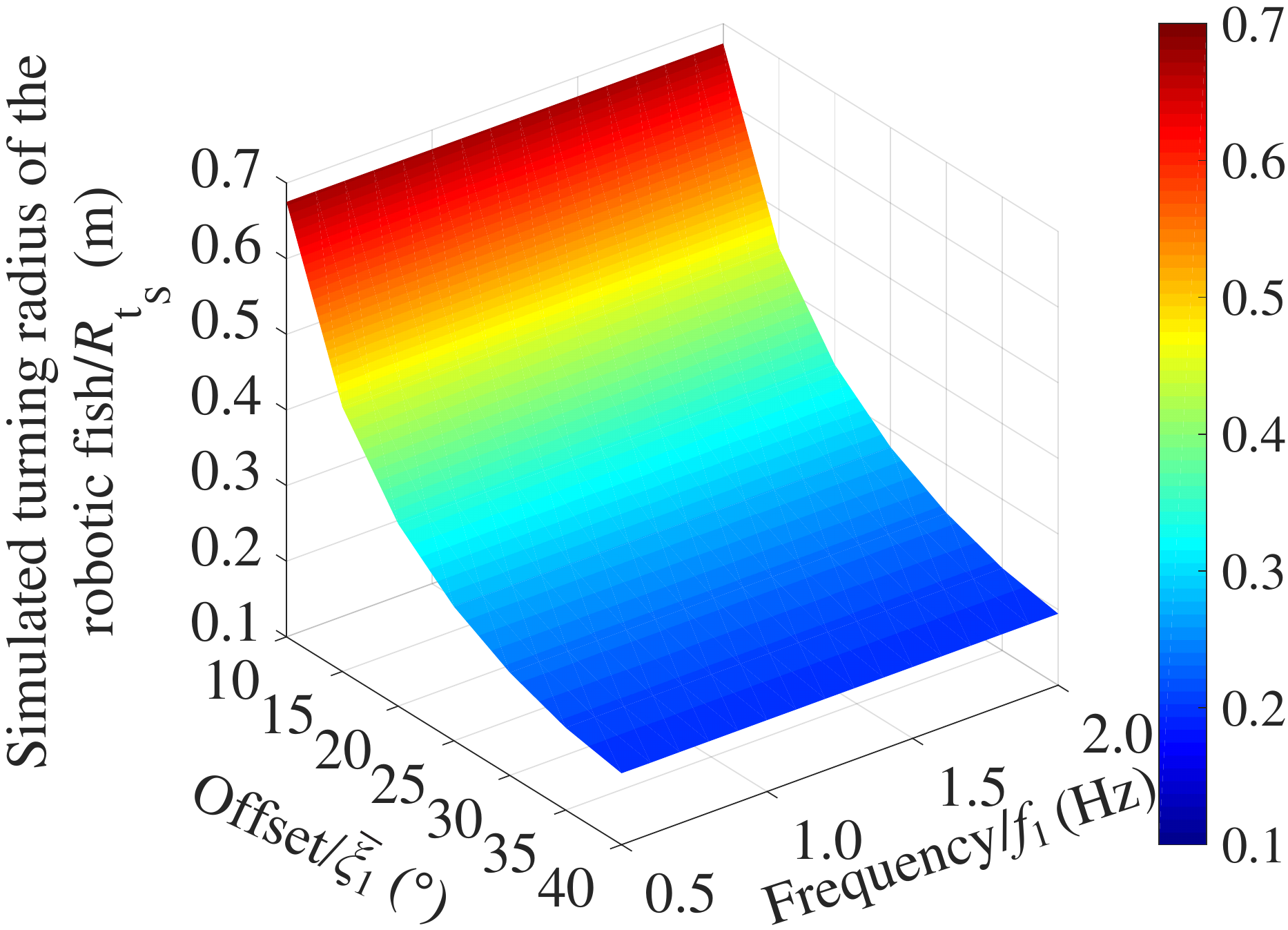}\\
(b)
\end{minipage}
\end{tabular}
\caption{Measured and simulated turning radius of the robotic fish. (a) Measured value. (b) Simulated value.}
\label{Actual turning radius of the robotic fish}
\end{figure}

\begin{figure}[htb]
\centering
\includegraphics[width=0.83\linewidth]{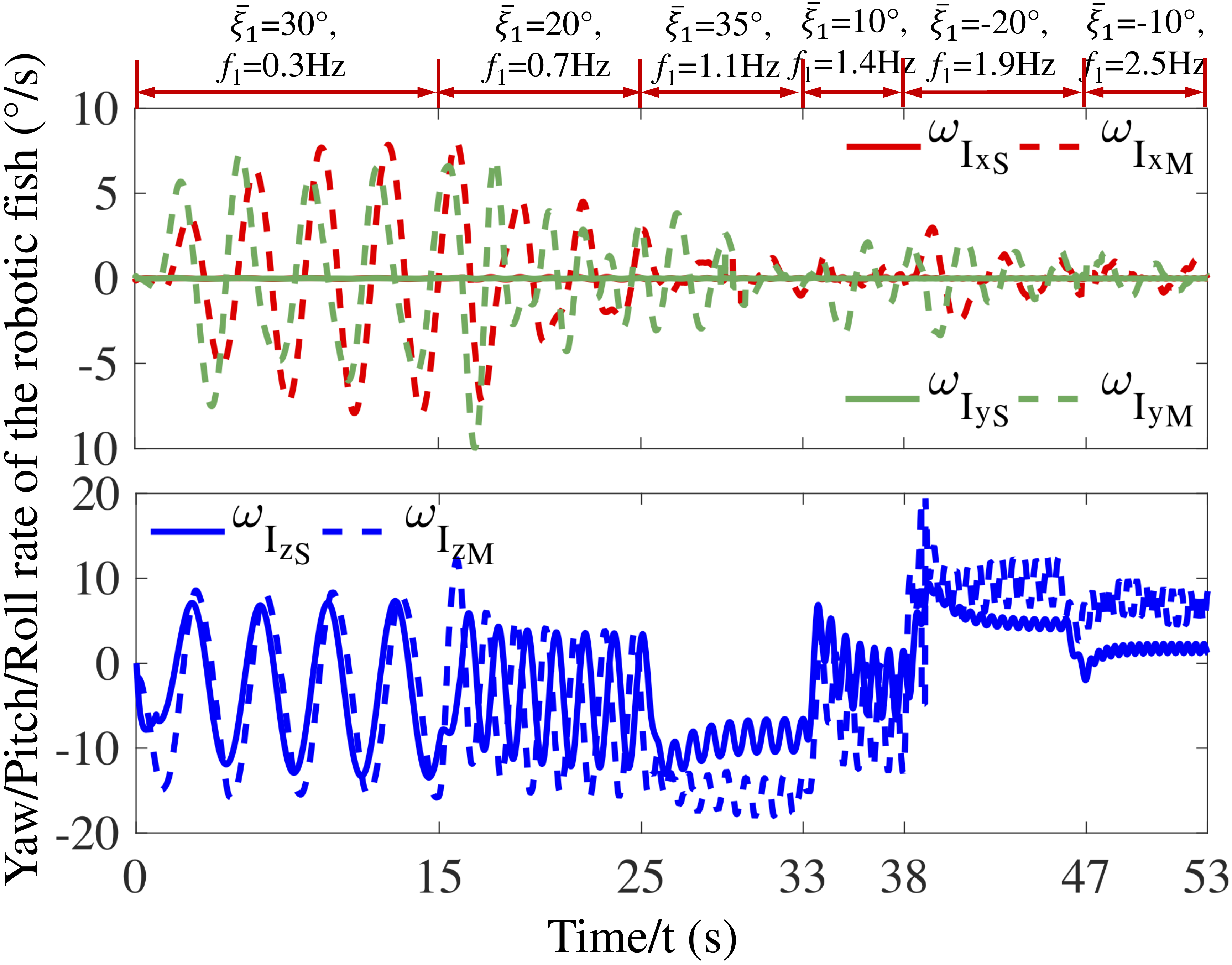}
\caption{Real-time yaw/pitch/roll rate of the robotic fish in turning motion under six combinations of $\bar{\xi_{1}}$ and $f_1$. $\omega_{I_{jS}}$ and $\omega_{I_{jM}} (j=x, y, z)$ indicate simulated and measured value of the yaw/pitch/roll rate, respectively.}
\label{Angular velocity of the robotic fish body}
\end{figure}

In turning motion experiment, varieties of turning angular velocities $\omega_t$ and turning radii $R_t$ were obtained by various combinations of oscillating offset $\bar{\xi_{1}}$ and frequency $f_1$ of the tail. As shown in Figure~\ref{Actual turning angular velocity of the robotic fish} and Figure~\ref{Actual turning radius of the robotic fish}, \revision{the measured value and simulated value} of $\omega_t$ match well with $R^2$=0.7462 and MAE=0.0409 rad/s, while the measured $R_t$ matches the simulated $R_t$ with a MAE=0.0657 m and an average percentage error of 18.5913\%. The $\omega_t$ increases with the increasing $\bar{\xi_{1}}$ and $f_1$. The $R_t$ decreases with the increasing $\bar{\xi_{1}}$ and it is nearly constant with the $f_1$. Figure~\ref{Angular velocity of the robotic fish body} shows the real-time yaw/pitch/roll rate of the robotic fish. It can \revisiontwo{be} seen that both the roll rate $\omega_{I_y}$ and pitch rate $\omega_{I_y}$ of the robotic fish oscillate around zero. The yaw rate $\omega_{I_z}$ oscillates around a positive value when the value of $\bar{\xi_{1}}$ is negative, in which case the robotic fish turns left. While the $\omega_{I_z}$ oscillates around a negative value when the value of $\bar{\xi_{1}}$ is positive, in which case the robotic fish turns right. A more careful inspection reveals that the amplitude of the $\omega_{I_z}$ increases with the $\bar{\xi_{1}}$ while decreases with the $f_1$\revisiontwo{, while} the rate of $\omega_{I_z}$ increases with the $f_1$. For the amplitudes of $\omega_{I_x}$ and $\omega_{I_y}$, they decrease with the $f_1$.
\subsection{Glidng motion}
\begin{figure}[htb]
\centering
\includegraphics[width=0.83\linewidth]{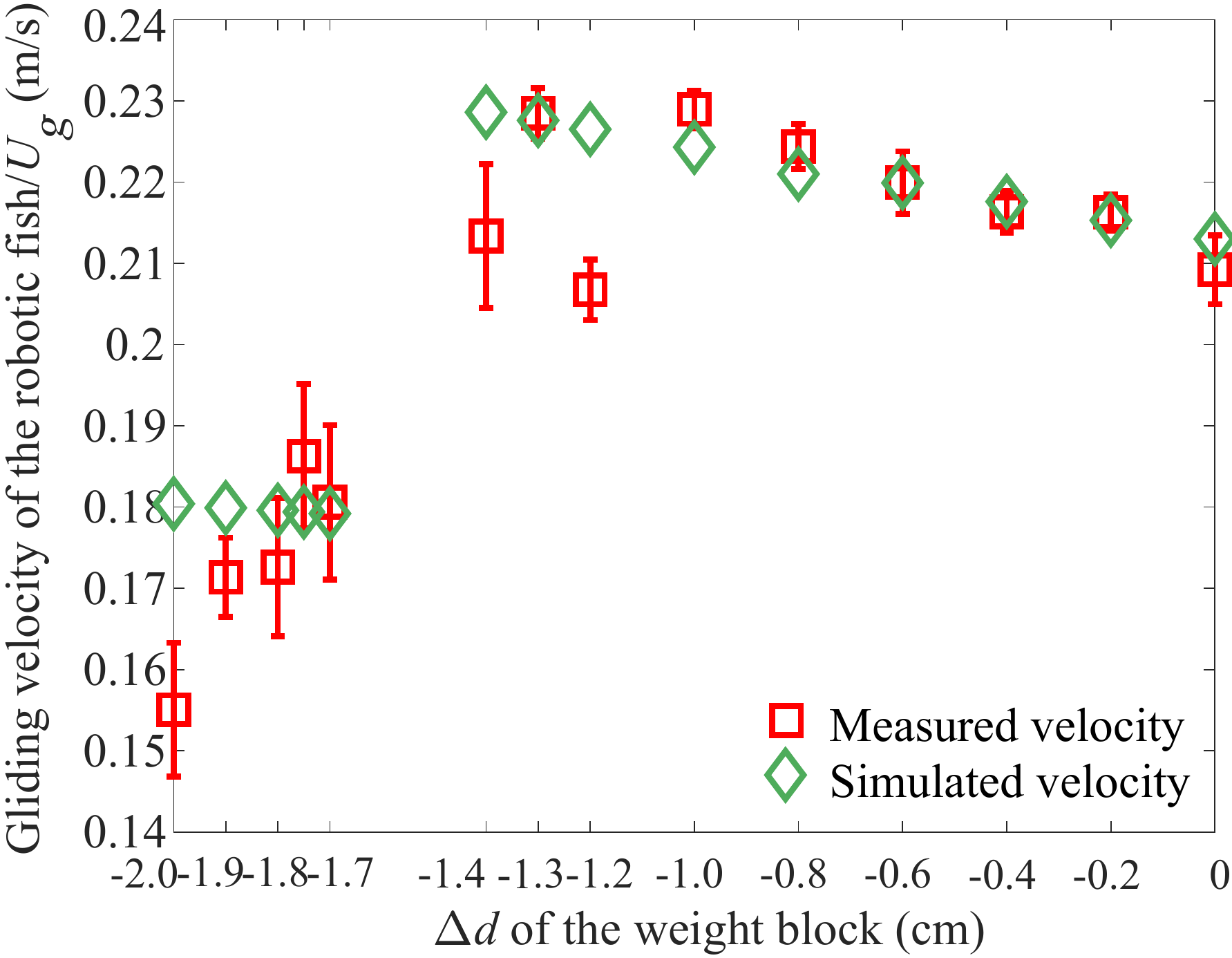}\\
\caption{Measured and simulated gliding velocity of the robotic fish.}
\label{Measured and simulated speed of the robotic fish in gliding motion}
\end{figure}
In glidng motion experiment, varieties of gliding velocities $U_g$ were obtained by changing $\Delta d$ of the weight block. $A_1$, $f_1$, and $\bar{\xi_{1}}$ of the tail are $20^\circ$, 2.0 Hz, and 0, respectively. Figure~\ref{Measured and simulated speed of the robotic fish in gliding motion} shows the measured and simulated $U_g$ of the robotic fish. The maximum and average percentage errors between the measured and simulated $U_g$ are 14.0507\% and 3.5340\%, respectively. It is noteworthy that because of the depth limitation of the water tank (only 0.8 m), the robotic fish reached the surface of the water before it reached the state of uniform motion. So the $U_g$ of the robotic fish for $\Delta d$=-2.0 cm, -1.9 cm, -1.8 cm, and -1.7 cm is average gliding velocity of the robotic fish in its acceleration process. While the $U_g$ for $\Delta d$ varied from -1.4 cm to 0 are velocities when the robotic fish was in uniform motion state. Comparing the $U_g$ for $\Delta d$ from -1.4cm to 0, it can be seen that $U_g$ of the robotic fish decreases with the increasing $\left | \Delta d \right |$.
\subsection{Spiral motion}
\begin{figure}[htb]
\centering
\includegraphics[width=0.83\linewidth]{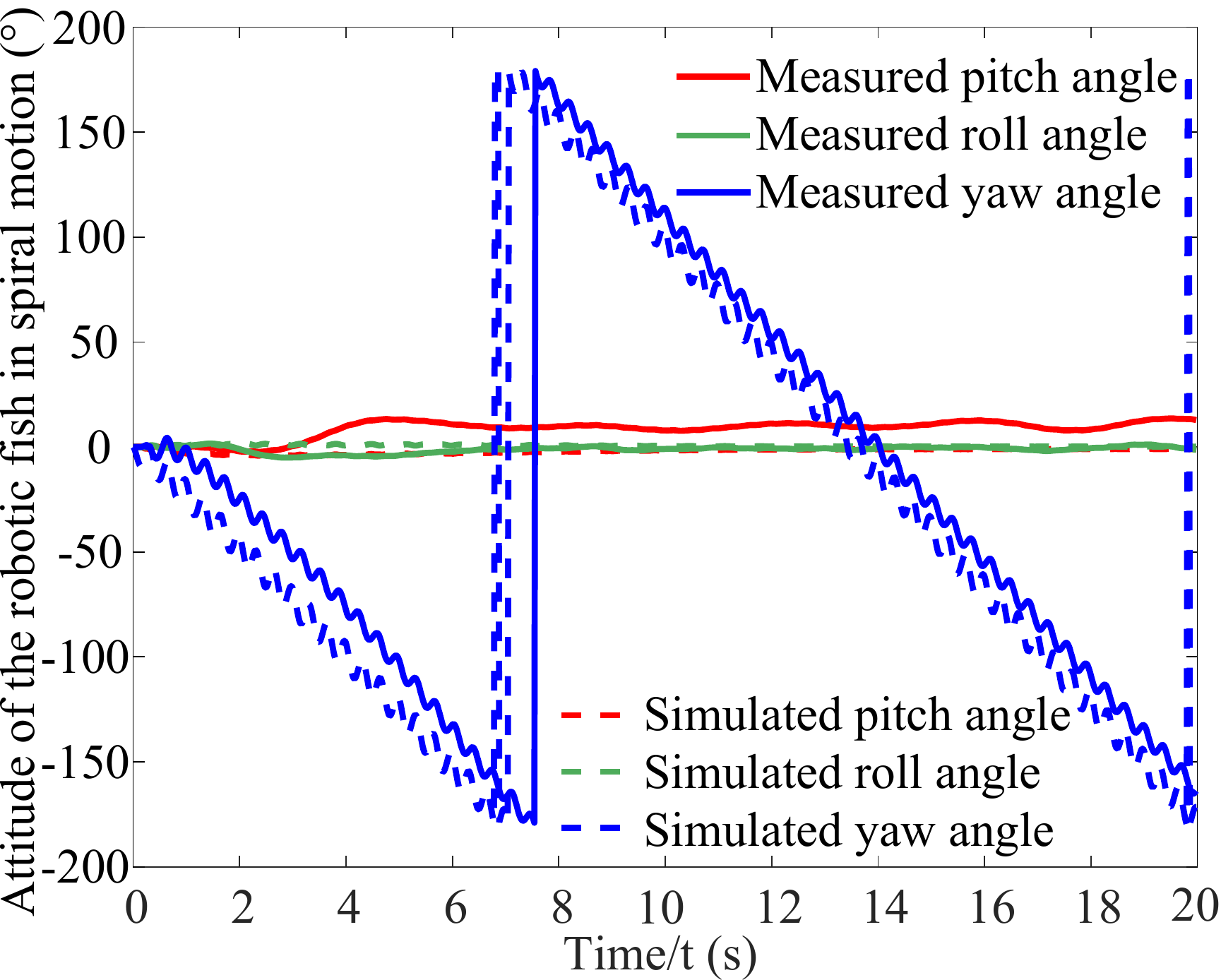}\\
\caption{Real-time measured and simulated attitude of the robotic fish in spiral motion.}
\label{Measured and simulated attitude of the robotic fish in spiral motion}
\end{figure}
\begin{figure}[htb]
\centering
\includegraphics[width=0.83\linewidth]{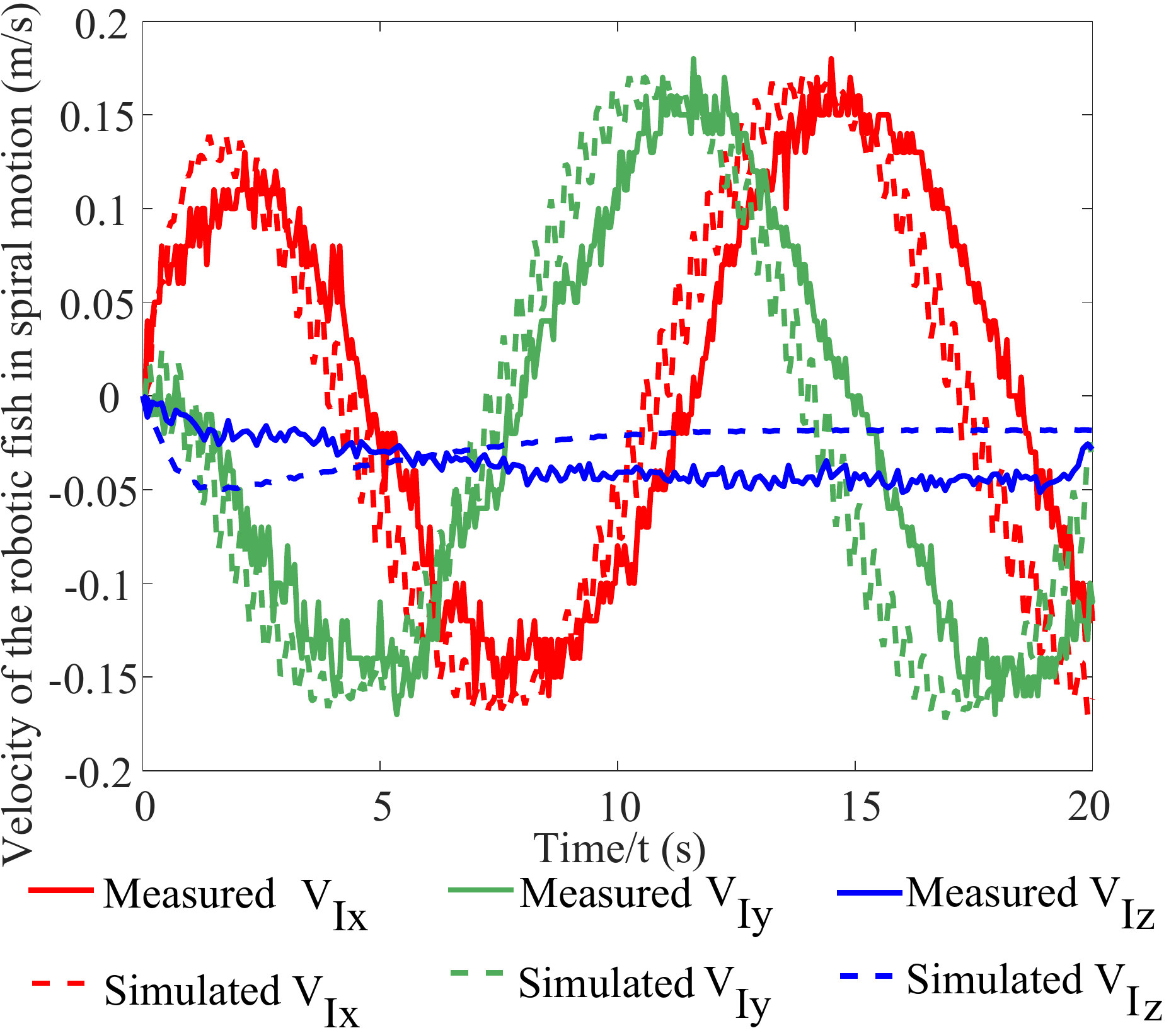}\\
\caption{Real-time velocity of the robotic fish in spiral motion.}
\label{Velocity of the robotic fish in spiral motion}
\end{figure}
\begin{figure}[htb]
\centering
\includegraphics[width=0.83\linewidth]{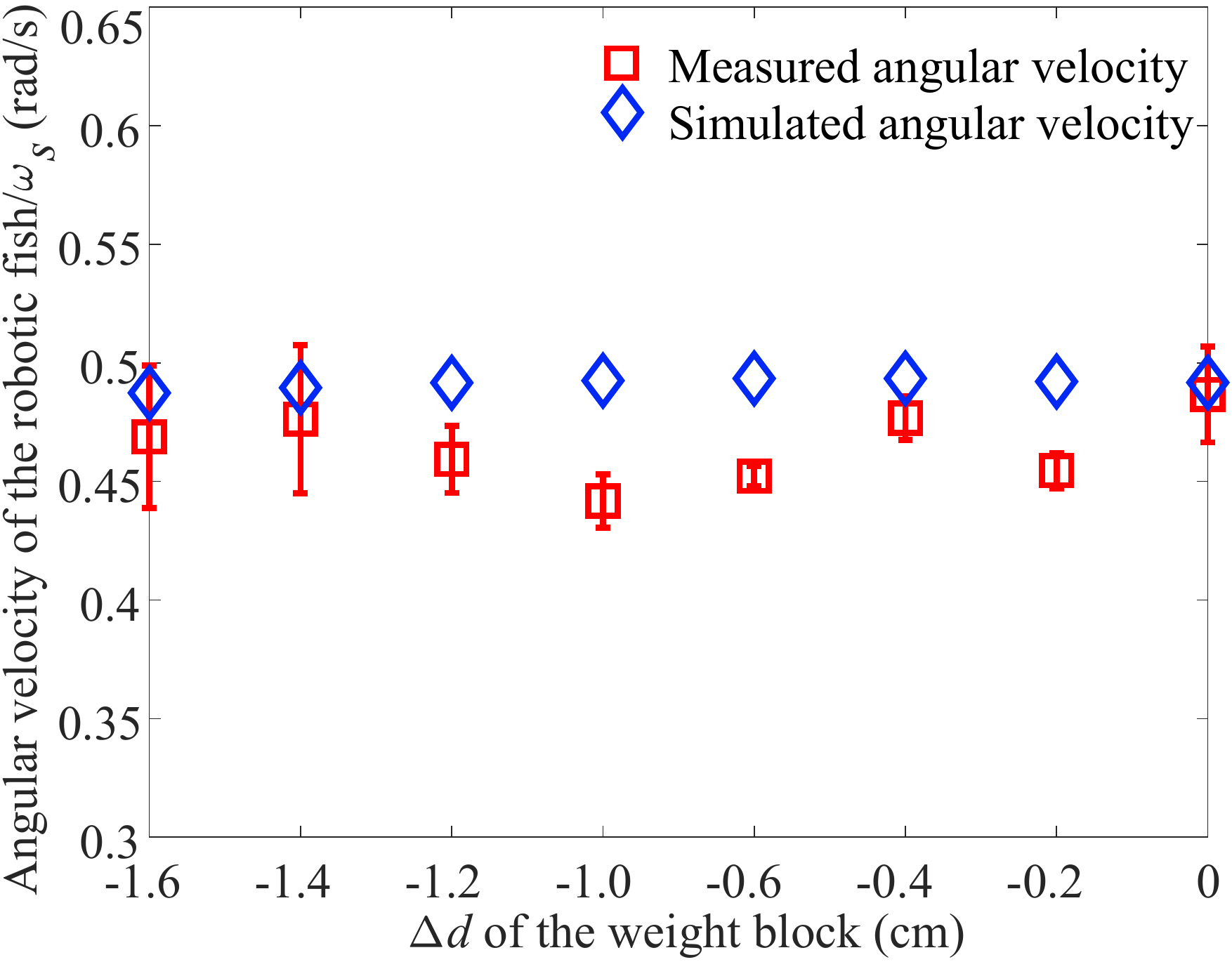}\\
\caption{Measured and simulated spiral angular velocity of the robotic fish.}
\label{Measured and simulated angular velocity of the robotic fish in spiral motion}
\end{figure}
\begin{figure}[htb]
\centering
\includegraphics[width=0.83\linewidth]{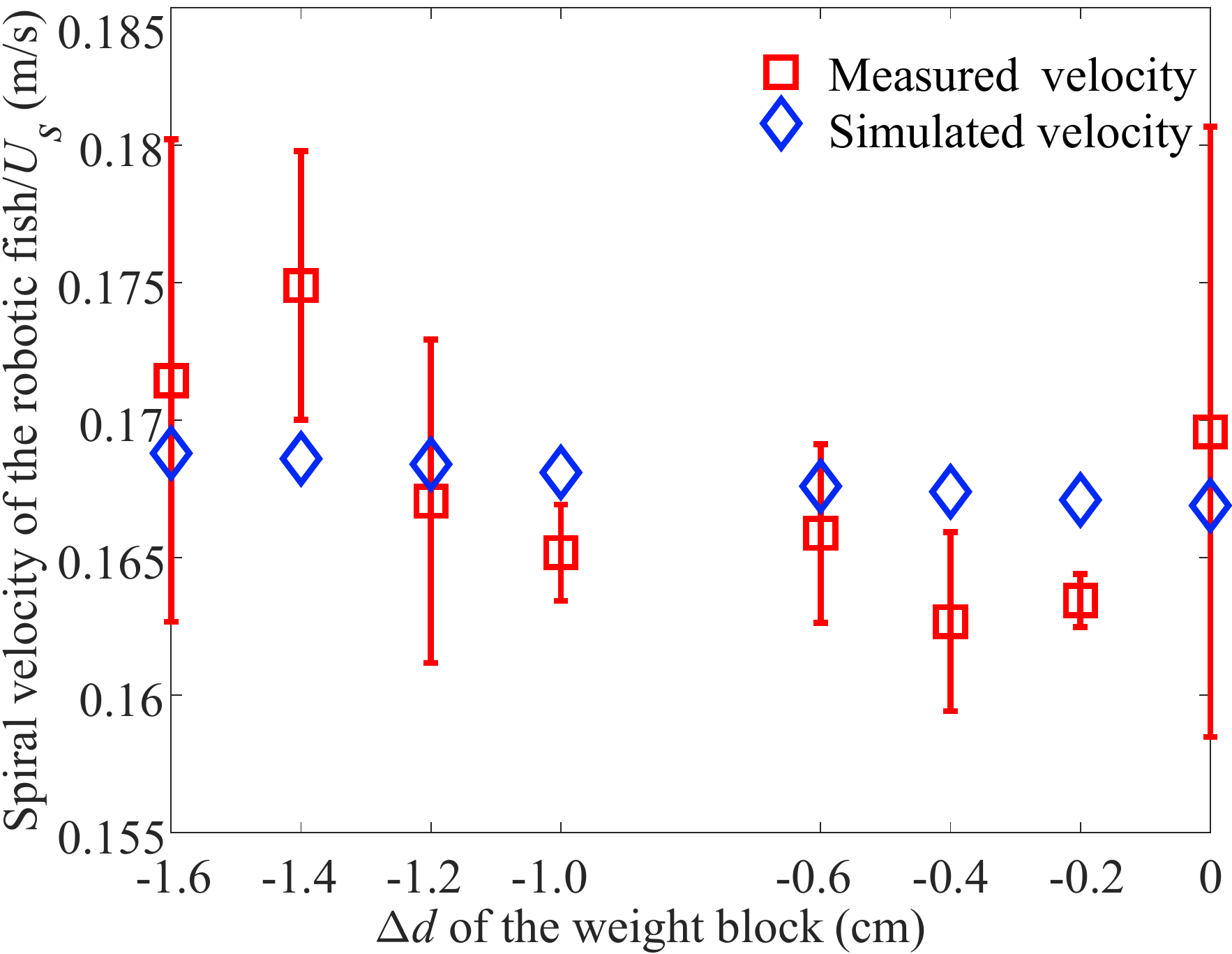}\\
\caption{Measured and simulated spiral velocity of the robotic fish.}
\label{Measured and simulated speed of the robotic fish in spiral motion}
\end{figure}
\begin{figure}[htb]
\centering
\includegraphics[width=0.83\linewidth]{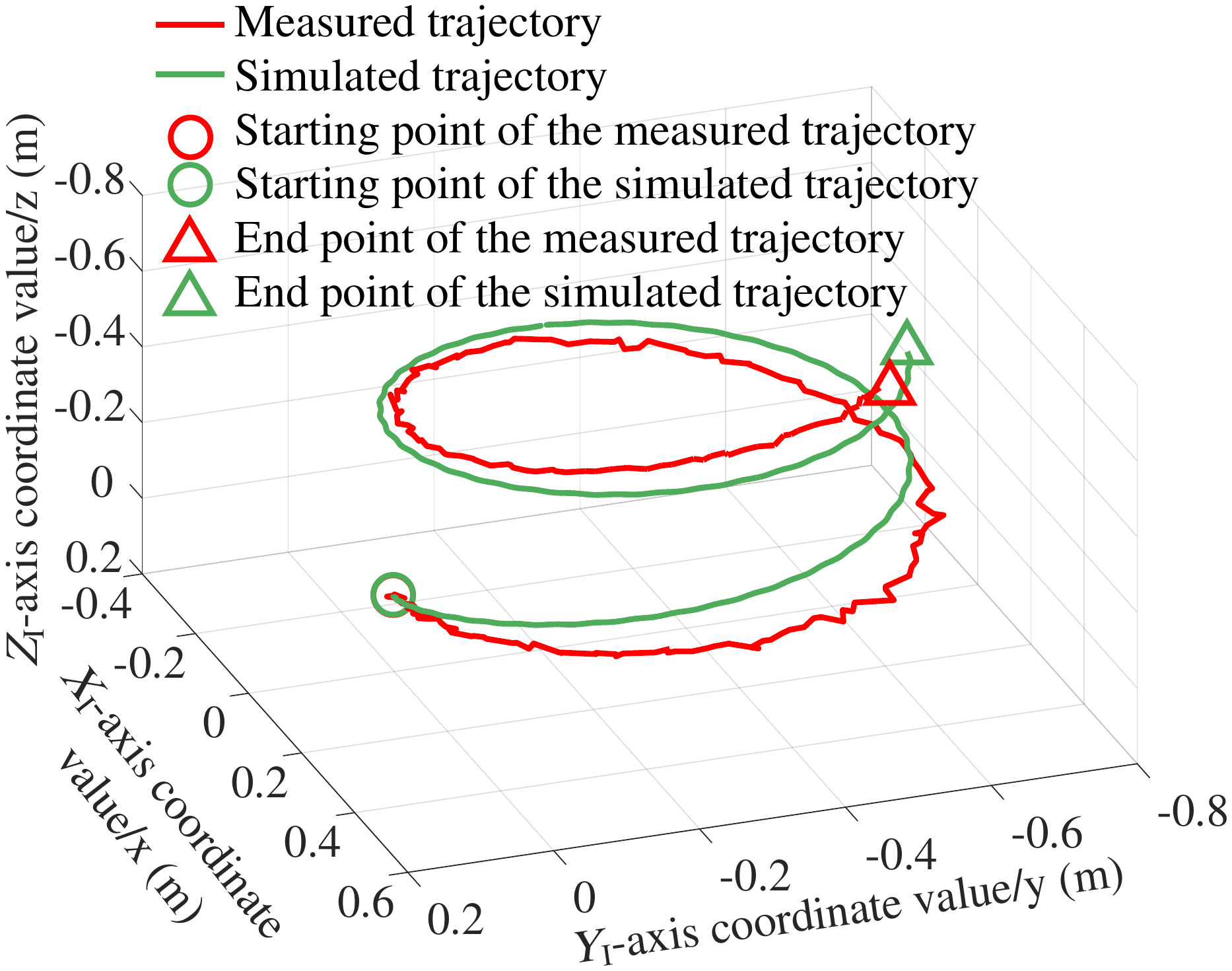}\\
\caption{Spiral trajectory of the robotic fish in spiral motion.}
\label{Trajectory of the robotic fish in spiral motion}
\end{figure}
The spiral motion was the result of a combination of non-zero $\Delta d$ and non-zero oscillating offset $\bar{\xi_{1}}$ of the tail. $A_1$ and $f_1$ of the tail are $20^\circ$ and 3.0 Hz, respectively. As shown in Figure~\ref{Measured and simulated attitude of the robotic fish in spiral motion}, yaw angle of the robotic fish oscillates around varied values with the time, while pitch angle and the roll angle oscillate around constant values. It can be seen that the simulated attitudes closely track the measured attitudes. The velocity $V_{I_x}$ along the axis $O_IX_I$ and the velocity $V_{I_y}$ along the axis $O_IX_I$ exhibit sine-like characteristics. The velocity $V_{I_z}$ along the axis $O_IZ_I$ gradually researches a negative value, which means the robotic fish is spiralling up. Figure~\ref{Measured and simulated angular velocity of the robotic fish in spiral motion} and Figure~\ref{Measured and simulated speed of the robotic fish in spiral motion} shows the measured and simulated spiral angular velocity $\omega_s$ and spiral velocity $U_s$ of the robotic fish, respectively. It can be seen that both the $\omega_s$ and the $U_s$ barely change with the $\Delta d$. The maximum and average percentage error of the $\omega_s$ are 3.6004\% and 1.9323\%, respectively. The maximum and average percentage error of the $U_s$ are 11.4808\% and 5.8953\%, respectively. Figure~\ref{Trajectory of the robotic fish in spiral motion} shows the measured and simulated spiral trajectory of the robotic fish in spiral motion. The measured trajectory tracks the simulated trajectory well with a maximum error of 0.3974 m.
\section{Conclusion and future work}
In this article, a dynamic model \revisiontwo{that} accounts for multiple three-dimensional motions, including rectilinear motion, turning motion, gliding motion, and spiral motion, of an active-tail-actuated robotic fish with barycentre regulating mechanism was proposed basing on Newton-Euler method. CAD software SolidWorks, HyperFlow based computational fluid dynamics (CFD) simulation, and grey-box model estimation method are used for determining model parameters. Massive kinematic experiments \revision{with robotic fish prototype} and numerical simulations demonstrate that the proposed model is capable of evaluating the trajectory, attitudes, and motion parameters including the linear velocity, motion radius, angular velocity, etc., for the robotic fish with small errors.

We are conducting researches on evaluating motion parameters of the robotic fish using its onboard ALLS, and an evaluation model \revisiontwo{that} links the linear velocity, angular velocity, and motion radius to the hydrodynamic pressure variations (PVs) surrounding the fish body has been preliminarily acquired. Using the PVs measured by the ALLS, the above-mentioned motion parameters can be evaluated by solving the evaluation model inversely. In the future work, we will input the ALLS-evaluated motion parameters into a dynamic \revisiontwo{model-based} controller as feedback terms, for adjusting the oscillation parameters of the robotic fish, and finally realizing flow-aided closed-loop control for the trajectory of the robotic fish.
\section*{Acknowledgment}
This work was supported in part by grants from the National Natural Science Foundation of China (NSFC, No. 91648120, 61633002, 51575005) and the Beijing Natural Science Foundation (No. 4192026).

\ifCLASSOPTIONcaptionsoff
  \newpage
\fi
\bibliographystyle{IEEEtran}
\bibliography{dynamic_and_hydrodynamic_modeling}

\begin{IEEEbiography}[{\includegraphics[width=1in,height=1.25in,clip,keepaspectratio]{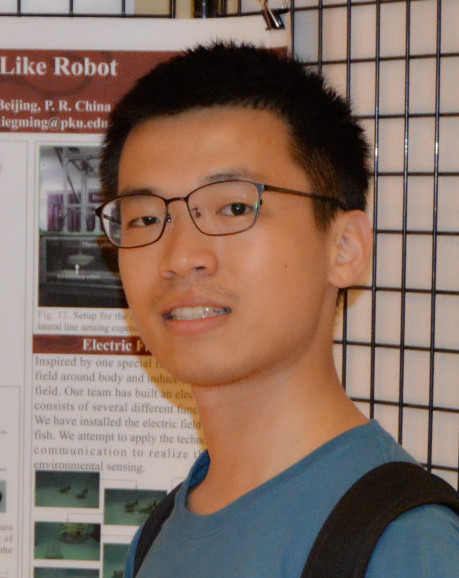}}]{Xingwen Zheng}
received the B.E. in Mechanical Engineering and Automation from Northeastern University, Shenyang, China in 2015. He is currently a PhD candidate at the Intelligent Biomimetic Design Lab, State Key Laboratory for Turbulence and Complex Systems, College of Engineering, Peking University, Beijing, China. His current research interests include biomimetic robotics, lateral line inspired sensing, and multi-robot control.
\end{IEEEbiography}
\begin{IEEEbiography}[{\includegraphics[width=1in,height=1.25in,clip,keepaspectratio]{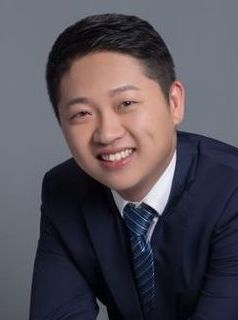}}]{Minglei Xiong}
received the B.E. in North China Electric Power University (NCEPU), Beijing, China in 2012. He is currently a PhD candidate in General Mechanics and Foundation of Mechanics at Peking University. His current research interests include biomimetic robotics, artificial intelligent, and game theory.
\end{IEEEbiography}
\begin{IEEEbiography}[{\includegraphics[width=1in,height=1.25in,clip,keepaspectratio]{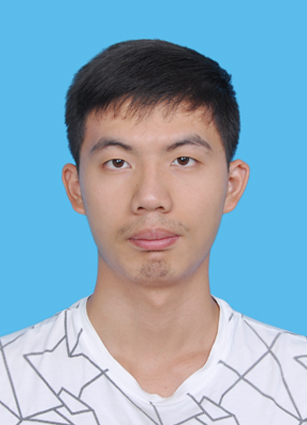}}]{Junzheng Zheng}
received the B.E. in Mechanical Design, Manufacturing and Automation from Huazhong University of Science and Technology, Wuhan, China in 2017. He is currently pursuing a master's degree in Control Theory and Control Engineering at Peking University. His current research interests include biomimetic robotics and lateral line inspired sensing.
\end{IEEEbiography}
\begin{IEEEbiography}[{\includegraphics[width=1in,height=1.25in,clip,keepaspectratio]{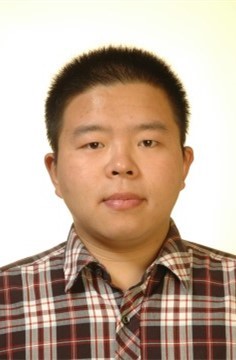}}]{Manyi Wang}
received the B.E. in Measurement and Control Technology and Instrumentation from National University of Defense Technology, Changsha, China in 2009. He is currently pursuing a master's degree in Control Theory and Control Engineering at Peking University. His current research interests include biomimetic robotics and lateral line inspired sensing.
\end{IEEEbiography}
\begin{IEEEbiography}[{\includegraphics[width=1in,height=1.25in,clip,keepaspectratio]{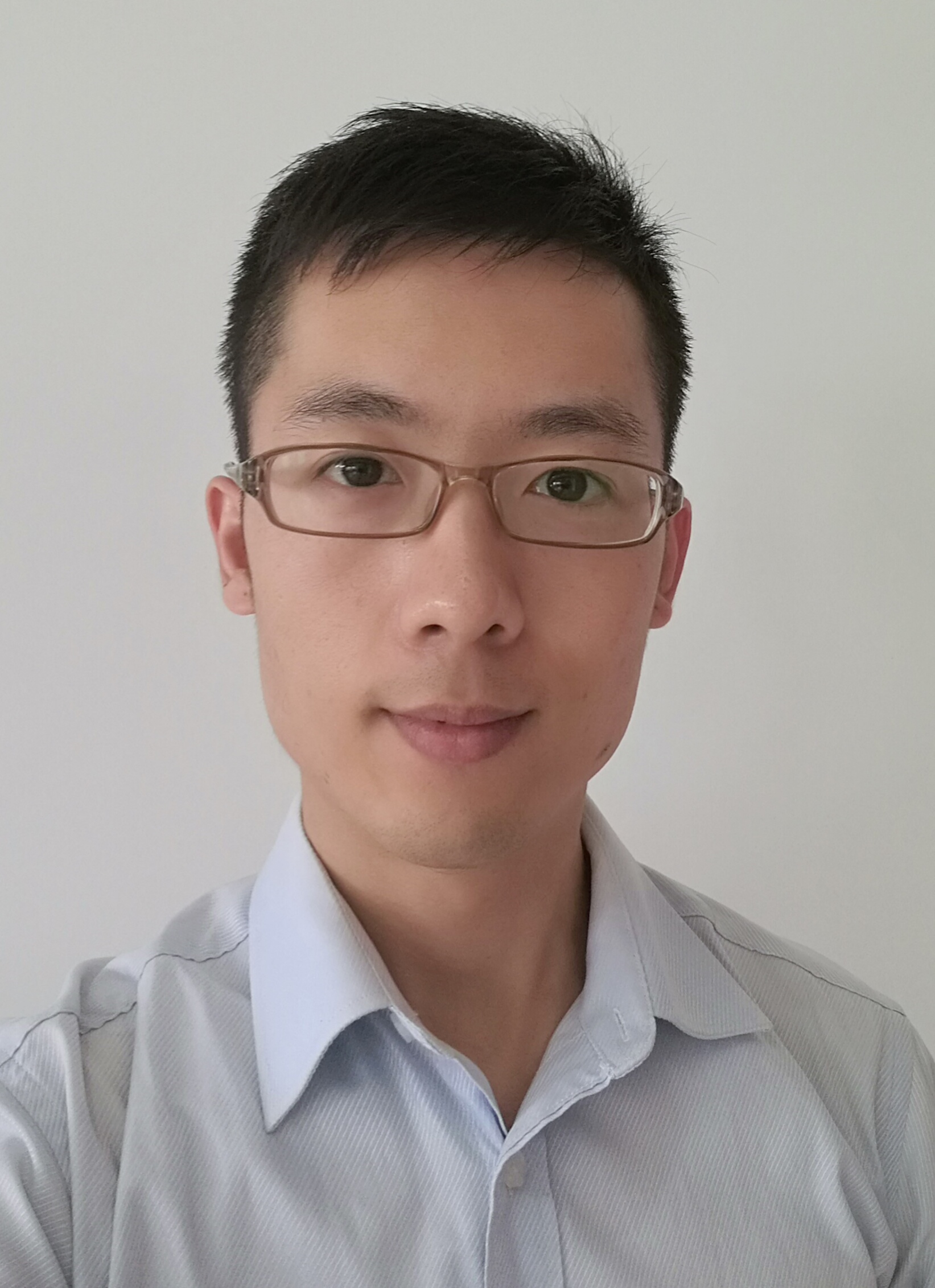}}]{Runyu Tian}
received the B.E. in Aircraft System and Engineering from National University of Defense Technology, Changsha, China in 2011. He is currently pursuing a master's degree in Control Theory and Control Engineering at Peking University. His current research interests include computational fluid dynamics, biomimetic robotics, and deep reinforcement learning.
\end{IEEEbiography}
\begin{IEEEbiography}[{\includegraphics[width=1in,height=1.25in,clip,keepaspectratio]{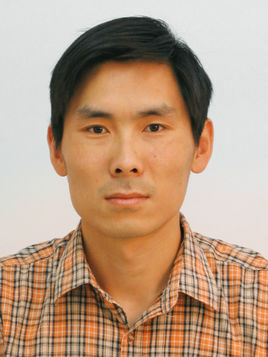}}]{Guangming Xie}
received his B.S. degrees in both Applied Mathematics and Electronic and Computer Technology, his M.E. degree in Control Theory and Control Engineering, and his Ph.D. degree in Control Theory and Control Engineering from Tsinghua University, Beijing, China in 1996, 1998, and 2001, respectively. Then he worked as a postdoctoral research fellow in the Center for Systems and Control, Department of Mechanics and Engineering Science, Peking University, Beijing, China from July 2001 to June 2003. In July 2003, he joined the Center as a lecturer. Now he is a Full Professor of Dynamics and Control in the College of Engineering, Peking University.

He is an Associate Editor of Scientific Reports, International Journal of Advanced Robotic Systems, Mathematical Problems in Engineering and an Editorial Board Member of Journal of Information and Systems Science. His research interests include smart swarm theory, multi-agent systems, multi-robot cooperation, biomimetic robot, switched and hybrid systems, and networked control systems.
\end{IEEEbiography}
\end{document}